\documentclass[10pt,twocolumn,letterpaper]{article}

\usepackage[pagenumbers]{cvpr} 

\usepackage{color, colortbl}
\usepackage{mathtools}
\usepackage{amsthm}
\usepackage{pifont}
\usepackage{bm}

\definecolor{Gray}{gray}{0.9}
\newcolumntype{g}{>{\columncolor{Gray}}c}

\DeclareMathOperator*{\argmin}{arg\,min}
\renewcommand{\eqref}[1]{Eq.~(\ref{#1})}

\makeatletter
\newcommand*{\rom}[1]{\expandafter\@slowromancap\romannumeral #1@}
\makeatother

\DeclareMathOperator{\ETC}{ETC}

\DeclareMathOperator{\SNR}{SNR}

\DeclareMathOperator{\VP}{VP}

\newcommand{\uu}{\bm{u}}
\newcommand{\xx}{\bm{x}}
\newcommand{\yy}{\bm{y}}

\newcommand{\ww}{\bm{w}}
\newcommand{\vv}{\bm{v}}

\newcommand{\abar}{\bar{\alpha}}
\newcommand{\hxx}{\hat{\bm{x}}}
\newcommand{\hep}{\hat{\bm{\epsilon}}}

\newcommand{\NN}{\mathcal{N}}

\newcommand{\EE}{\mathbb{E}}
\newcommand{\RR}{\mathbb{R}}

\newcommand{\PP}{\mathbb{P}}
\newcommand{\QQ}{\mathbb{Q}}

\newcommand{\cmark}{\textcolor{ForestGreen}{\ding{51}}}
\newcommand{\xmark}{\textcolor{red}{\ding{55}}}

\newtheorem{proposition}{Proposition}

\newcommand{\add}[1] {\textcolor{black}{#1}} 

\definecolor{cvprblue}{rgb}{0.21,0.49,0.74}
\usepackage[pagebackref,breaklinks,colorlinks,allcolors=cvprblue]{hyperref}
\usepackage{algorithm, algorithmicx, algpseudocode, tablefootnote}

\def\paperID{12377} 
\def\confName{CVPR}
\def\confYear{2025}

\title{Latent Schr\"{o}dinger Bridge: Prompting Latent Diffusion for Fast Unpaired Image-to-Image Translation}

\author{
Jeongsol Kim$^{* \, 1}$ \quad Beomsu Kim$^{* \, 2}$ \quad Jong Chul Ye$^2$ \\
$^*$Equal contribution \\
$^1$Department of Bio and Brain Engineering, KAIST \\
$^2$Kim Jaechul Graduate School of AI, KAIST \\
{\tt\small \{jeongsol, beomsu.kim, jong.ye\}@kaist.ac.kr}}

\begin{document}
\twocolumn[{%
\renewcommand\twocolumn[1][]{#1}%
\maketitle
\begin{center}
    \centering
    \captionsetup{type=figure}
    \includegraphics[width=0.8\linewidth]{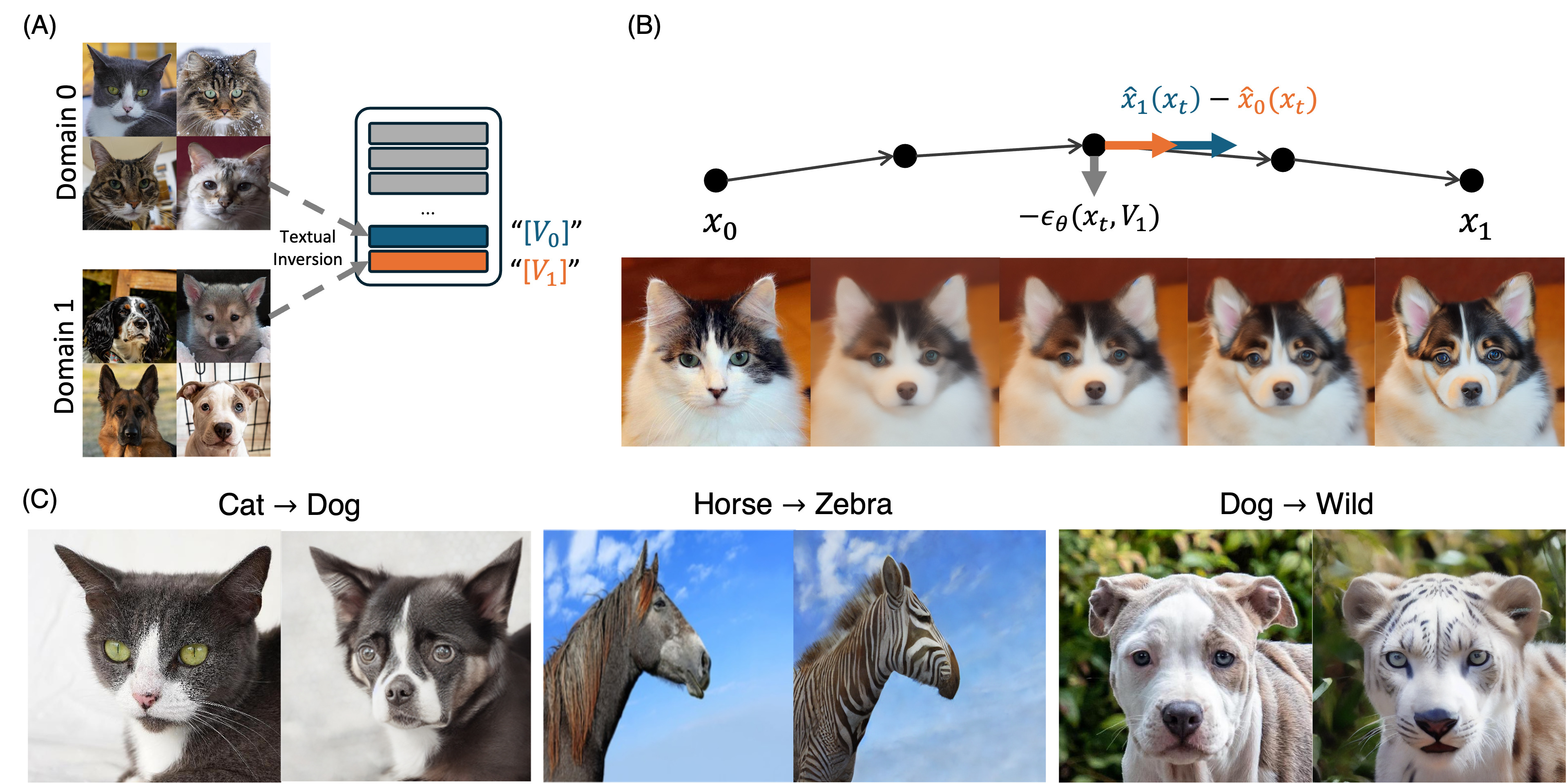}
\caption{\textbf{Overview of the proposed method}. (A) Using an unpaired dataset, we optimize text embeddings to obtain conditioned score functions using Stable Diffusion 1.5. (B) By decomposing Schrödinger Bridge (LSB) ODE with these optimized score functions, (C) we achieve high-quality image-to-image translation with as few as 8 NFEs.}
\label{fig:overview}
\end{center}}]

\begin{abstract}
Diffusion models (DMs), which enable both image generation from noise and inversion from data, have inspired powerful unpaired image-to-image (I2I) translation algorithms. However, they often require a larger number of neural function evaluations (NFEs), limiting their practical applicability. In this paper, we tackle this problem with Schr\"{o}dinger Bridges (SBs), which are stochastic differential equations (SDEs) between distributions with minimal transport cost. We analyze the probability flow ordinary differential equation (ODE) formulation of SBs, and observe that we can decompose its vector field into a linear combination of source predictor, target predictor, and noise predictor. Inspired by this observation, we propose Latent Schr\"{o}dinger Bridges (LSBs) that approximate the SB ODE via pre-trained Stable Diffusion, and develop appropriate prompt optimization and change of variables formula to match the training and inference between distributions. We demonstrate that our algorithm successfully conduct competitive I2I translation in unsupervised setting with only a fraction of computation cost required by previous DM-based I2I methods.
\end{abstract}

\begin{figure*}[t]
    \centering
    \includegraphics[width=0.85\linewidth]{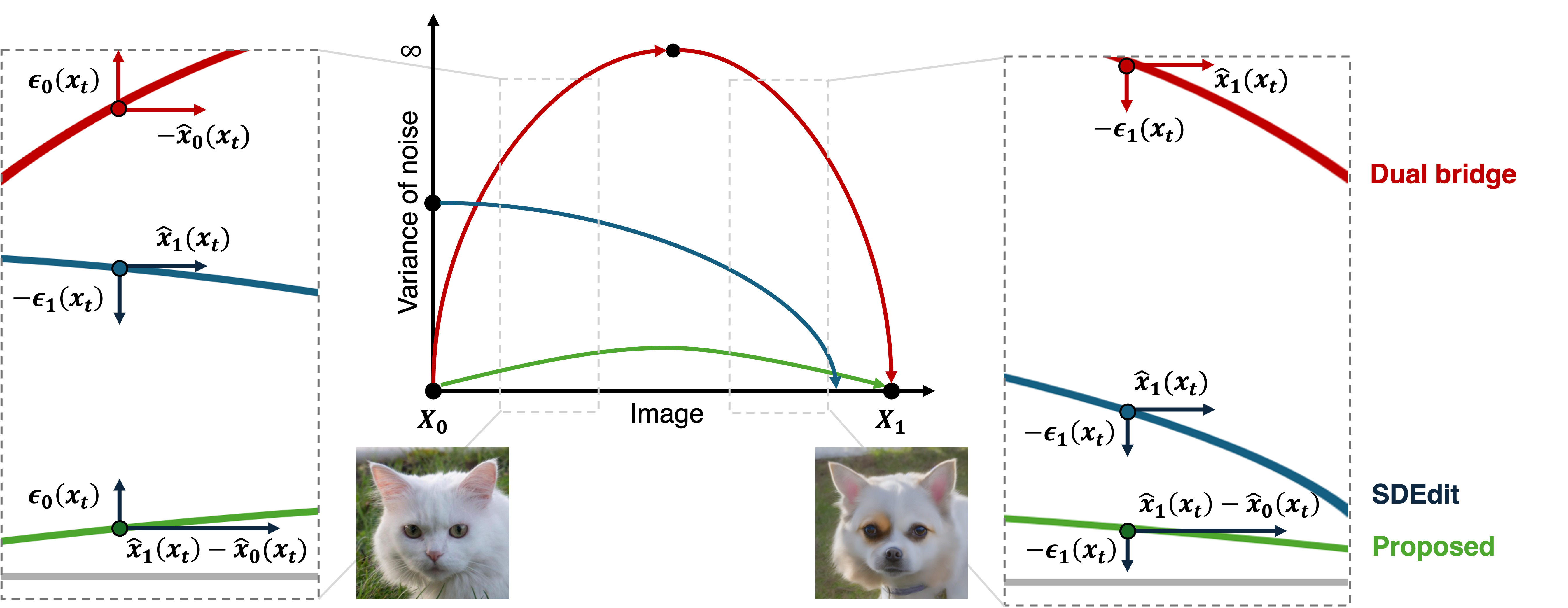}
    \caption{Decomposition of different ODEs for image to image translation. \add{Dual Bridge requires inversion to Gaussian noise, causing large errors in fast translation; SDEdit lacks a repelling force from the source domain, resulting in incomplete translation. In contrast, the LSB ODE avoids inversion and includes a repelling term from the source image, enabling effective translation to the target image}.}
    \label{fig:concept}
    \vspace{-2mm}
\end{figure*}

\section{Introduction} \label{sec:intro}

Diffusion models (DMs) \citep{song2021sde,ho2020ddpm,song2021ddim}, which learn to generate data from noise by iterative denoising, achieves state-of-the-art results in a wide variety of generative learning tasks such as unconditional generation \citep{dhariwal2021beat}, solving inverse problems \citep{chung2023dps}, text-to-image generation \citep{rombach2022ldm}, and so on. A particular trait of DMs is that they enable exact inversion from data to latent noise by solving the probability flow ordinary differential equation (ODE) \citep{song2021sde}. This has motivated numerous image-to-image (I2I) translation methods which translate between two domains by concatenating two diffusion models \citep{su2023ddib,tumanyan2023pnp}. Such methods often outperform traditional GAN-based translation methods thanks to powerful diffusion priors.

However, diffusion models (DMs) are often slow, requiring large number of neural function evaluations (NFEs) for high-quality image synthesis. This drawback has inspired numerous accelerated diffusion sampling methods. While there are plenty of works which explore sampling acceleration for unconditional generation such as fast solvers \citep{song2021ddim,lu2022dpmsolver,zhang2023deis,zhou2024amed} or distillation \citep{song2023cm,song2024icm}, fast sampling for diffusion-based image-to-image (I2I) translation remain relatively under-explored.

Schr\"{o}dinger Bridges (SBs), which learn stochastic differential equations (SDEs) between two arbitrary distributions, is a promising alternative to diffusion-based I2I translation \citep{wang2021deepsb,bortoli2021dsb,shi2023dsbm}. Specifically, SBs are optimal in the sense that they minimize entropy-regularized transport cost, and thus enabling fast translation. Though they are attractive in theory, exact SB training often requires orders of magnitude longer training than DMs.

To mitigate this problem, we propose a sampling procedure inspired by Schr\"{o}dinger Bridges (SBs), which leverage pre-trained DMs to approximate SB probability flow ODEs. 
Specifically, we find that the SB probability flow ODE velocity can be expressed as a linear combination of a source domain predictor, a target domain predictor, and a noise predictor. This opens up possibilities for leveraging pre-trained latent diffusion model to approximate SB probability flow ODEs.
More specifically, we obtain source (resp. target) domain predictors using pretrained score functions in Stable Diffusion (SD) by simply
optimizing the text prompt using source (resp. target) domain example data.
The noise predictor can be also computed similarly.
Furthermore, based on the observation that naively plugging in SB probability flow ODE states into pre-trained scores does not work due to SNR mismatch between training and inference stage inputs, we derive a change of variables formula to align the SNR for diffusion and SB variables.
The resulting algorithm, termed the Latent Schrödinger Bridge (LSB), requires minimal additional complexity 
 yet outperforms numerous DM-based image-to-image (I2I) methods with significantly fewer NFEs. We demonstrate that our algorithm is highly scalable and surpasses diffusion-based I2I translation across a wide range of datasets.
%

\begin{figure*}[t]
    \centering
    \includegraphics[width=0.8\linewidth]{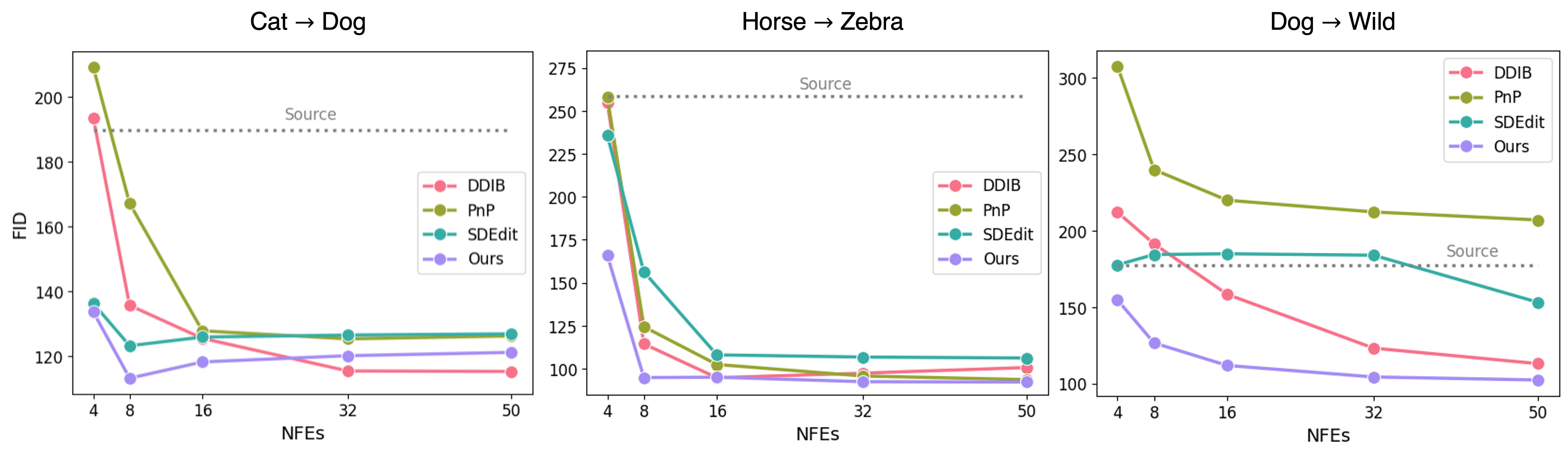}
    \caption{FID vs. NFE for three image translation tasks. \add{We evaluate FID on translated images from baseline methods with various NFEs. For small NFE $\leq 10$, LSB ODE outperforms baselines and the quality is improved or maintained with more NFEs.}}
    \label{fig:nfe}
    \vspace{-3mm}
\end{figure*}

\section{Related Works} \label{sec:related}

\textbf{Unpaired image-to-image (I2I) translation.} Given two or more image domains, the goal of I2I translation is to translate images from one domain to another while maintaining input image structure. Early works such as Pix2Pix \citep{isola2017pix2pix} worked with paired datasets that consist of structurally similar image pairs, \ie, input-label pairs, and train Generative Adversarial Networks (GANs) \citep{goodfellow2014gan} to maximize pixel-wise similarity between translated inputs and labels. However, such strategy is not applicable to unpaired datasets, which have motivated works such as cycle-consistent GANs \citep{zhu2017cyclegan}, 
Contrastive Unsupervised Translation \citep{park2020cut}, etc., which regularize GANs with some weaker notion of similarity between inputs and translated outputs. More recent works such as SDEdit \citep{meng2022sdedit} or Dual Diffusion Implicit Bridge (DDIB) \citep{su2023ddib} leverage powerful diffusion priors to translate images from one domain to another by corrupting the input image with Gaussian noise, then denoising with target domain diffusion model.

\noindent
\textbf{Fast diffusion sampling.} While diffusion models (DMs) possess powerful generative performance, they often require a large number of neural function evaluations (NFEs) to generate data from noise. This is because DMs translate Gaussian noise to data by integrating stochastic differential equations (SDEs) or ordinary differential equations (ODEs), so naively reducing NFE increases truncation error and degrades sample quality \citep{song2021sde}. This has motivated works on diffusion distillation \citep{song2023cm,song2024icm,salimans2022pd} or fast solvers \citep{song2021ddim,lu2022dpmsolver,zhang2023deis,zhou2024amed} for few-step generation. While such methods show strong performance on unconditional generation, there is a lack of works on fast I2I translation via DMs.

\noindent
\textbf{Schr\"{o}dinger bridges (SBs).} SBs are also a class of neural SDEs, but unlike DMs, SBs can translate samples between arbitrary domains with minimal transport cost. There are roughly three branches of algorithms for learning SBs. Flow Matching \citep{lipman2023fm,tong2023cfm} methods distill SB between minibatches using minibatch optimal transport. Iterative Proportional Fitting (IPF) methods \citep{bortoli2021dsb,shi2023dsbm,liu2024gsbm} iteratively minimize transport cost by training models on input-output pairs generative by itself. Finally, adversarial approaches \citep{kim2024unsb,gushchin2024asbm} leverage a discriminator to enforce distributional constraints while solving the SB problem.
In the following, we provide more detailed description of SBs.

\section{Background} \label{sec:background}

\noindent
\textbf{Mathematical formulation.}
Let $\PP_0$ and $\PP_1$ be two data distributions on $\RR^d$. Schr\"{o}dinger Bridges (SBs) are Stochastic Differential Equations (SDEs) of the form
\begin{align}
    d\xx_t = \uu(\xx_t,t) \, dt + \sqrt{\tau} d\ww_t \label{eq:sb_sde}
\end{align}
on $t \in [0,1]$ whose drift $\uu$ minimizes transport cost
\begin{align}
    \min_{\uu} \, \EE\left[ \int_0^1 \frac{1}{2} \| \uu_\theta(\xx_t,t) \|^2 \, dt \right] \label{eq:scp}
\end{align}
subject to the boundary conditions $\xx_i \sim \PP_i$ for $i \in \{0,1\}$. Here, the expectation is taken respect to trajectories sampled from \eqref{eq:sb_sde}, and $\tau$ is an hyper-parameter which controls the amount of noise in the SB trajectories.

SBs have multiple practical benefits over DMs. First, SBs do not require $\PP_0$ or $\PP_1$ to be Gaussian, enabling unpaired image-to-image (I2I) translation. Furthermore, SBs enable fast generation, since they are constrained to minimize transport cost per \eqref{eq:scp}. Indeed, in the noiseless limit $\tau \rightarrow 0$, the SB SDE becomes an optimal transport (OT) ODE, which enables translation between $\PP_0$ and $\PP_1$ with a single Euler discretization step.

Theoretically, SBs transport samples between $\PP_0$ and $\PP_1$ according to entropy-regularized OT. Concretely, given
\begin{align}
    \begin{cases}
        \PP_{01}^\tau \coloneqq \argmin_{\Gamma_{01} \in \Pi(\PP_0,\PP_1)} \ETC(\Gamma_{01};\tau) \\
        d\PP_{t|01}^\tau(\xx_t | \xx_0,\xx_1) \coloneqq \NN(\xx_t | \bm{\mu}_t, \sigma_t^2 \bm{I})
    \end{cases} \label{eq:eot}
\end{align}
where entropy-regularized transport cost is defined as
\begin{align*}
    \ETC(\Gamma_{01};\tau) \coloneqq \EE_{(\xx_0,\xx_1) \sim \Gamma_{01}} [ \|\xx_0 - \xx_1\|_2^2 ] - 2 \tau H(\Gamma_{01})
\end{align*}
and conditional mean and variance are given as
\begin{align*}
    \bm{\mu}_t \coloneqq (1-t) \xx_0 + t \xx_1, &\quad
    \sigma_t^2 \coloneqq t(1-t)\tau,
\end{align*}
the variable $\xx_t$ of the SDE \eqref{eq:sb_sde} at time $t$ is distributed according to $\PP_t^\tau$ \citep{tong2023cfm}. Here, $\Pi(\PP_0,\PP_1)$ is the collection of joint distributions with marginals $\PP_0$ and $\PP_1$, and $H$ denotes information entropy.


Assuming we have access to $\PP_{01}^\tau$, it is then possible to characterize the SB probability flow ODE, \ie, the deterministic counterpart of the SB SDE, as a solution to a regression problem \citep{tong2023cfm}. Specifically, given the following vector field conditioned on endpoints $(\xx_0,\xx_1)$,
\begin{align*}
    \textstyle \uu(\xx_t | \xx_0, \xx_1, t) \coloneqq \frac{(1/2-t)}{t(1-t)}(\xx_t - \bm{\mu}_t) + (\xx_1 - \xx_0),
\end{align*}
if $\vv$ solves the regression problem
\begin{align}
    \min_{\vv} \EE_{(\xx_0,\xx_1,\xx_t) \sim \PP_{01t}^\tau} [\| \uu(\xx_t | \xx_0,\xx_1,t) - \vv(\xx_t,t) \|_2^2] \label{eq:sb_cfm}
\end{align}
then the ODE
\begin{align}
    d\xx_t = \vv(\xx_t,t) \, dt \label{eq:sb_pfode}
\end{align}
generates the SB probability path $\PP_t^\tau$, \ie, samples $\xx_t$ that follow \eqref{eq:sb_pfode} are distributed according to $\PP_t^\tau$ at time $t$.


\noindent
\textbf{Limitations of previous SB methods.} 
As discussed in Section \ref{sec:related}, algorithms for learning SB can be roughly grouped into three branches: Conditional Flow Matching (CFM), Iterative Proportional Fitting (IPF), and adversarial learning. The first approach, CFM, uses entropic OT maps between minibatches of $\PP_0$ and $\PP_1$ as a proxy for the global OT map $\PP_{01}^\tau$ in \eqref{eq:eot}. Due to its reliance on minibatches, it is vulnerable to the curse of dimensionality, especially when the training dataset is high-dimensional or sparse (see Section 3 in \citep{kim2024unsb}). IPF methods require training upon synthetic data, which causes error to accumulate in the marginals. Finally, adversarial approaches require training of an auxiliary discriminator, which is known to be unstable.

Given the current status of SB methods, it is natural to ask whether we can simulate SBs efficiently, even when given a sparse set of unpaired source and target domain images. In the next section, we propose Latent SB, which mitigates both problems by approximating the SB probability flow ODE velocity $\vv$ with pretrained latent diffusion models such as Stable Diffusion. As we shall show through experiments in Section \ref{sec:experiments}, our formulation allows us to exploit rich prior information in Stable Diffusion model to overcome the curse of dimensionality with minimal amount of training.

\section{Latent Schr\"{o}dinger Bridge} \label{sec:method}

In the following three sections, we construct Latent Schr\"{o}dinger Bridge (LSB) ODE for fast unpaired image-to-image (I2I) translation. Concretely, in Section \ref{sec:decomp}, we show that the velocity for the Schr\"{o}dinger Bridge (SB) probability flow ordinary differential equation (ODE) can be decomposed into three interpretable terms. In Section \ref{sec:snr_matching}, we derive change-of-variables formulae that allow us to approximate each term with diffusion score functions. In Section \ref{sec:practical}, we introduce practical techniques for implementing the proposed method using Stable Diffusion.

\subsection{SB ODE Vector Field Decomposition} \label{sec:decomp}

The solution to \eqref{eq:sb_cfm} is given as
\begin{align}
    \vv(\xx_t,t) = \EE_{(\xx_0,\xx_1) \sim \PP_{01|t}^\tau(\cdot,\cdot | \xx_t)}[\uu(\xx_t | \xx_0,\xx_1,t)]
\end{align}
which, by linearity of expectation, can be decomposed into
\begin{align}\label{eq:sb_cfm_sol}
    \textstyle \vv(\xx_t,t) = \frac{(1/2-t)}{t(1-t)} \EE[(\xx_t - \bm{\mu}_t)] + \left( \EE[\xx_1] - \EE[\xx_0] \right)
\end{align}
where we have dropped the subscripts on expectations for notational brevity. We now simplify each expectation.

The second expectation can be expressed as
\begin{align}
    \EE_{(\xx_0,\xx_1) \sim \PP_{01|t}^\tau(\cdot,\cdot | \xx_t)}[\xx_1] = \EE_{\xx_1 \sim \PP_{1|t}^\tau(\cdot | \xx_t)}[\xx_1]
    \label{eq:x0hat}
\end{align}
which is essentially a prediction of the target image given current state $\xx_t$, so we shall denote it as $\hxx_1^\tau(\xx_t)$. It acts as an attracting force towards the target domain.
Analogously, the third expectation is
\begin{align}
    \EE_{(\xx_0,\xx_1) \sim \PP_{01|t}^\tau(\cdot,\cdot | \xx_t)}[\xx_0] = \EE_{\xx_0 \sim \PP_{0|t}^\tau(\cdot | \xx_t)}[\xx_0]
    \label{eq:x1hat}
\end{align}
which is a prediction of the source image given $\xx_t$, so we shall denote it as $\hxx_0^\tau(\xx_t)$. Due to its negative sign in \eqref{eq:sb_cfm_sol}, it acts as a repelling force from the source domain.
As for the first expectation, using the reparametrization trick \citep{kingma2014vae}, samples $\xx_t$ from $\PP_{t|01}^\tau$ can be expressed as
\begin{align*}
    \xx_t = \bm{\mu}_t + \sigma_t \bm{\epsilon}
\end{align*}
where $\bm{\epsilon} \sim \NN(\bm{0},\bm{I})$, from which it follows that
\begin{align}
    \begin{cases}
        d\PP^\tau_{01\bm{\epsilon}}(\xx_0,\xx_1,\bm{\epsilon}) = d\PP^\tau_{01}(\xx_0,\xx_1) \NN(\bm{\epsilon} | \bm{0}, \bm{I}), \\
        d\PP^\tau_{t|01\bm{\epsilon}}(\xx_t | \xx_0,\xx_1,\bm{\epsilon}) = \delta_{\bm{\mu}_t + \sigma_t \bm{\epsilon}}(\xx_t).
    \end{cases} \label{eq:sb_joint_cond}
\end{align}
Hence
\begin{align}
    &\EE_{(\xx_0,\xx_1) \sim \PP_{01|t}^\tau(\cdot,\cdot | \xx_t)}[(\xx_t - \bm{\mu}_t)] \notag \\
    &= \xx_t - \EE_{(\xx_0,\xx_1) \sim \PP_{01|t}^\tau(\cdot,\cdot | \xx_t)}[\bm{\mu}_t] \notag \\
    &= \xx_t - \EE_{(\xx_0,\xx_1,\bm{\epsilon}) \sim \PP_{01\bm{\epsilon}|t}^\tau(\cdot,\cdot,\cdot | \xx_t)}[\bm{\mu}_t + (\xx_t - \bm{\mu}_t - \sigma_t \bm{\epsilon})] \notag \\
    &= \sigma_t \cdot \EE_{(\xx_0,\xx_1,\bm{\epsilon}) \sim \PP_{01\bm{\epsilon}|t}^\tau(\cdot,\cdot,\cdot | \xx_t)}[ \bm{\epsilon} ] \notag \\
    &= \sigma_t \cdot \EE_{\bm{\epsilon} \sim \PP_{\bm{\epsilon}|t}^\tau(\cdot | \xx_t)}[ \bm{\epsilon} ] \label{eq:eps_predictor}
\end{align}
where at the third line, we have used the fact that
\begin{align*}
    d\PP^\tau_{\bm{\epsilon}|01t}(\bm{\epsilon} | \xx_0,\xx_1,\xx_t) = \delta_{(\xx_t-\bm{\mu}_t)/\sigma_t}(\bm{\epsilon})
\end{align*}
and so
\begin{align*}
    \EE_{(\xx_0,\xx_1,\bm{\epsilon}) \sim \PP_{01\bm{\epsilon}|t}^\tau(\cdot,\cdot,\cdot | \xx_t)}[\xx_t - \bm{\mu}_t - \sigma_t \bm{\epsilon}] = \bm{0}.
\end{align*}
\eqref{eq:eps_predictor} is essentially a prediction of Gaussian noise added to $\bm{\mu}_t$, so we shall denote the expectation in \eqref{eq:eps_predictor} concisely as $\hep^\tau(\xx_t)$. Its sign in the velocity \eqref{eq:sb_cfm_sol} is positive when $t < 0.5$ and negative when $t > 0.5$. Thus, the SB ODE adds noise in the first half of the unit interval, and removes noise in the second half.
Putting everything together,
\begin{align}
     \vv(\xx_t,t) &= \textstyle \frac{(1/2-t)}{t(1-t)} \cdot \sigma_t \cdot \hep^\tau(\xx_t) + \hxx_1^\tau(\xx_t) - \hxx_0^\tau(\xx_t) \notag \\
    &= \textstyle \frac{(1/2-t)\sqrt{\tau}}{\sqrt{t(1-t)}} \cdot \hep^\tau(\xx_t) + \hxx_1^\tau(\xx_t) - \hxx_0^\tau(\xx_t) \label{eq:decomp_vel}
\end{align}
which shows that the SB probability flow ODE velocity is a linear combination of three simple terms: noise prediction, target image prediction, and source image prediction.

\paragraph{Using General Predictors.} 
Recall that the predictors $\hxx_0^\tau$, $\hxx_1^\tau$, $\hep^\tau$ are posterior means of $\xx_0$, $\xx_1$, $\bm{\epsilon}$ with respect to the joint $\PP_{01\bm{\epsilon}}^\tau$ and the conditional $\PP_{t|01\bm{\epsilon}}^\tau$ defined in \eqref{eq:sb_joint_cond}. \add{However, learning $\PP_{01}^\tau$ (the solution to entropy-regularized OT problem \eqref{eq:eot}) for finite values of $\tau$ is known to be notoriously difficult, so we ask what happens if we replace $\PP_{01}^\tau$ with an arbitrary $\Gamma_{01} \in \Pi(\PP_0,\PP_1)$, \ie, define}
\begin{align}
    \begin{cases}
        d\PP_{01\bm{\epsilon}}(\xx_0,\xx_1,\bm{\epsilon}) = d\Gamma_{01}(\xx_0,\xx_1) \NN(\bm{\epsilon} | \bm{0}, \bm{I}), \\
        d\PP_{t|01\bm{\epsilon}}(\xx_t | \xx_0,\xx_1,\bm{\epsilon}) = \delta_{\bm{\mu}_t + \sigma_t \bm{\epsilon}}(\xx_t),
    \end{cases} \label{eq:gamma_joint_cond}
\end{align}
and use general predictors
\begin{align}
    \begin{cases}
        \hxx_0(\xx_t) = \EE_{\xx_0 \sim \PP_{0|t}(\cdot|\xx_t)}[\xx_0] \\
        \hxx_1(\xx_t) = \EE_{\xx_1 \sim \PP_{1|t}(\cdot|\xx_t)}[\xx_1] \\
        \hep(\xx_t) = \EE_{\bm{\epsilon} \sim \PP_{\bm{\epsilon}|t}(\cdot|\xx_t)}[\bm{\epsilon}]
    \end{cases} \label{eq:gamma_preds}
\end{align}
in place of $\hxx_0^\tau$, $\hxx_1^\tau$, $\hep^\tau$ to simulate the SB ODE velocity \eqref{eq:sb_cfm_sol}. Interestingly, as shown by the following proposition, we can use any coupling $\Gamma_{01} \in \Pi(\PP_0,\PP_1)$, and \eqref{eq:sb_pfode} would still be a valid probability flow ODE between the marginals $\PP_0$, $\PP_1$.

\begin{proposition} \label{prop:gamma}
    The ODE with velocity
    \begin{align}
        \textstyle \vv(\xx_t,t) = \frac{(1/2-t)\sqrt{\tau}}{\sqrt{t(1-t)}} \cdot \hep(\xx_t) + \hxx_1(\xx_t) - \hxx_0(\xx_t) \label{eq:gamma_decomp_vel}
    \end{align}
    defined with predictors in \eqref{eq:gamma_preds} translates samples from $\PP_0$ to $\PP_1$ when solved from $t = 0$ to $1$, and vice versa when solved from $t = 1$ to $0$.
\end{proposition}

In fact, in the special case when $\PP_1$ is standard normal, we obtain the diffusion probability flow ODE for certain hyper-parameter choices. The next proposition serves as a sanity check corroborating the correctness of Proposition \ref{prop:gamma}.

\begin{proposition} \label{prop:sanity}
    When $\tau = 0$, $\Gamma_{01} = \PP_0 \otimes \PP_1$, and $\PP_1$ is standard normal, the ODE with velocity \eqref{eq:gamma_decomp_vel} is equivalent to the diffusion probability flow ODE.
\end{proposition}

Intuitively speaking, Proposition \ref{prop:gamma} implies that as long as the predictors $\hxx_0$, $\hxx_1$, $\hep$ yield proper estimates of the source image, target image, and Gaussian noise given current state $\xx_t$, we can still leverage the SB ODE \eqref{eq:sb_pfode} to translate between domains. \add{While the generated marginals $\PP_t$ will no longer be exact SB marginals $\PP_t^\tau$ when $\Gamma_{01} \neq \PP_{01}^\tau$ for intermediate values of $t \in (0,1)$, we can interpret this as a trade-off between transport cost (see \eqref{eq:scp}) and predictor learnability.}

This observation motivates us to see whether diffusion models can be used to simulate source, target, and noise predictors. Specifically, we wish to simulate \eqref{eq:gamma_preds} with variance-preserving (VP) diffusion noise prediction models $\hep_{\VP}^0$ and $\hep_{\VP}^1$ trained on $\PP_0$ and $\PP_1$, resp.

\subsection{Matching VP and SB Signal-to-Noise Ratios} \label{sec:snr_matching}

Without loss of generality, let us consider how we may use $\hep^0_{\VP}$ to simulate source domain and noise predictors $\hxx_0$ and $\hep$. By time symmetry of SB ODEs, we can easily derive target domain and noise predictors $\hxx_1$ and $\hep$ with $\hep^1_{\VP}$. We first introduce some notation for describing VP diffusion.

Given $\xx_0 \sim \PP_0$ and $\bm{\epsilon} \sim \NN(\bm{0},\bm{I})$, VP noise prediction models $\hep^0_{\VP}$ learn to recover $\bm{\epsilon}$ given\footnote{\add{We denote the VP diffusion variable by $\yy_s$, defined for $s \in (0,\infty)$, to distinguish it from the SB variable $\xx_t$, defined for $t \in (0,1)$.}}
\begin{align}
    \yy_s = \sqrt{\abar_s} \xx_0 + \sqrt{1-\abar_s} \bm{\epsilon}
\end{align}
as an input, where $\abar_0 = 1$ and $\abar_s \rightarrow 0$ as $s \rightarrow \infty$. Or, theoretically speaking,
\begin{align}
    \hep^0_{\VP}(\yy_s) = \EE_{\bm{\epsilon} \sim \QQ_{\bm{\epsilon}|s}(\cdot | \yy_s)}[\bm{\epsilon}]
\end{align}
where the noise posterior is defined w.r.t.
\begin{align}
    \begin{cases}
        d\QQ_{0\bm{\epsilon}}(\yy_0,\bm{\epsilon}) = d\QQ_0(\yy_0) \NN(\bm{\epsilon} | \bm{0}, \bm{I}) \\
        d\QQ_{s | 0 \bm{\epsilon}}(\yy_s | \yy_0, \bm{\epsilon}) = \delta_{\sqrt{\abar_s} \yy_0 + \sqrt{1-\abar_s} \bm{\epsilon}}(\yy_s)
    \end{cases} \label{eq:vp_joint_cond}
\end{align}
for $\QQ_0 = \PP_0$. Hence, it is straightforward to obtain estimates for $\bm{\epsilon}$ and $\xx_0$ when we are given $\yy_s$:
\begin{align}
    \begin{cases}
        \hep(\yy_s) = \hep^0_{\VP}(\yy_s), \\
        \hxx_0(\yy_s) = (\yy_s - \sqrt{1 - \abar_s} \hep^0_{\VP}(\yy_s)) / \sqrt{\abar_s}.
    \end{cases} \label{eq:vp_pred}
\end{align}
We note that the above expression for $\hxx_0(\yy_s)$ is well-known as \textit{Tweedie's formula} for denoising \citep{kim2021tweedie}.

However, when we compare the expressions for $\xx_t$ in \eqref{eq:gamma_joint_cond} and $\yy_s$ in \eqref{eq:vp_joint_cond}, reproduced here for clarity,
\begin{align*}
    &\xx_t = \bm{\mu}_t + \sigma_t \bm{\epsilon}, \\
    &\yy_s = \sqrt{\abar_s} \xx_0 + \sqrt{1 - \abar_s} \bm{\epsilon},
\end{align*}
we observe that they have different formulations, so there is a distribution mismatch between inputs $\xx_t$ that we wish to plug into $\hep_{\VP}^0$ during SB ODE inference, and inputs $\yy_s$ to $\hep_{\VP}^0$ that are actually used during VP diffusion training.

The largest mismatch arises from the fact that $\xx_t$ and $\yy_s$ have different signal-to-noise ratios (SNRs), \ie, the ratio of coefficients multiplied to the signal component ($\bm{\mu}_t$ for $\xx_t$, and $\xx_0$ for $\yy_s$) and coefficients multiplied to the noise component ($\bm{\epsilon}$ for $\xx_t$ and $\yy_s$) are different. To address this discrepancy, we equate the SNR of $\xx_t$ and $\yy_s$, which yields the conversion formulae
\begin{align}
    \begin{cases}
        \abar_s = 1/(\sigma_t^2+1), \\
        \yy_s = \sqrt{\abar_s} \xx_t = \xx_t / \sqrt{\sigma_t^2 + 1},
    \end{cases}
    \label{eq:conversion}
\end{align}
that can now be plugged into \eqref{eq:vp_pred} to yield noise and $\xx_0$ predictions given $\xx_t$. \add{See Appendix \ref{append:proofs} for a full derivation.}

We remark that we can calculate $s$ by treating $\abar_s$ as a continuous function of $s$, and applying its inverse to $(\sigma_t^2 + 1)^{-1}$, \ie, $s = \abar^{-1}((\sigma_t^2 + 1)^{-1})$. Since the SB ODE is symmetrical with respect to $t = 0.5$, we can use identical conversion formulae to use $\bm{\epsilon}^1_{\VP}$ for $\xx_1$ and noise prediction. We now discuss practical considerations for implementing our method
using a pretrained Stable Diffusion model.

\subsection{Leveraging Stable Diffusion Models}
 \label{sec:practical}
To mitigate the increased computational cost of the proposed method, we employ a single pre-trained Stable Diffusion model to obtain estimates in both the source and target domains. This approach offers several advantages. First, implementing our SB decomposition in the latent domain reduces the dimensionality of samples $\xx_t$ in the latent domain and the associated diffusion models, thanks to the encoder's dimensionality reduction role in the SB model. Consequently, this reduces overall computational complexity. For simplicity, and with a slight abuse of notation, we use $\xx_t$ to denote the latent variable update.
Second, by assigning different text description for source and target domain through text prompt,
we can  condition a single diffusion model to compute predictors for both source and target domain.
However, a n\"{a}ive application of this approach results in suboptimal performance. In the following, we provide a detailed description of practical considerations.

\noindent\textbf{Prompt optimization.}
In this study, we assume an unsupervised setting, where paired datasets and textual descriptions for each domain are not available. For example, if we aim to translate dog images into ``wild"\footnote{We use the label name from the AFHQ dataset.} domain, which includes various wild feline species, it is insufficient to describe the target domain merely as ``wild animal".
Thus, we optimize the text embedding using a given set of images from a domain to condition a SD  model, as illustrated in Figure~\ref{fig:overview}. Specifically, we employ the textual inversion technique \citep{gal2022textual}, designed for personalized image generation. For further details, please refer the Appendix.
Empirically, we found that less than 1k images are enough to obtain text embeddings that leads to proper predictors of LSB ODE.


\noindent\textbf{ODE initial point.}
In the absence of target estimation $\xx_1$, we define the initial point for the proposed LSB ODE by
\begin{align}
    \xx_{t_0} = (1-t_0)\xx_0 + \sqrt{t_0(1-t_0)\tau}\bm{\epsilon}
    \label{eq:init}
\end{align}
where $t_0 \geq 0$ denotes initial SB time, $\tau$ denotes the variance of SB SDE (see Section \ref{sec:background}), and $\bm{\epsilon} \sim \mathcal{N}(\bm{0}, \bm{I})$.

\noindent
\textbf{Time-dependent noise predictor.} As noted in the previous section, we can use either $\hep^0_{\VP}$ or $\hep^1_{\VP}$ to predict noise from $\xx_t$. As $\xx_t$ is closer to the $\PP_0$ domain when $t < 0.5$ and $\PP_1$ domain when $t > 0.5$, we use $\hep^0_{\VP}$ when $t < 0.5$ and $\hep^1_{\VP}$ when $t > 0.5$ to improve noise estimates.

\noindent
\textbf{Classifier-free guidance (CFG).}
As observed in previous studies on diffusion models, 
we can improve the reflectance of text prompt by using the classifier free guidance when computing \eqref{eq:x0hat} and \eqref{eq:x1hat}. Specifically, with null-text embedding $\varnothing$,
\begin{align}
    \hat\xx_0^{(\omega)}(\xx_t) = (\xx_t - \sqrt{1-\bar\alpha_t}\bm{\epsilon}^{(\omega)}_\theta(\xx_t, c_0))/\sqrt{\bar\alpha_t}
\end{align}
where $\bm{\epsilon}^{(\omega)}_\theta(\xx_t, c_0) = (1-\omega)\bm{\epsilon}_\theta(\xx_t,\varnothing) + \omega \bm{\epsilon}_\theta(\xx_t,c_0)$ denotes predicted noise with CFG. This is equivalent to scaling $\xx_0$ and $\xx_1$ predictors by $\omega$:
\begin{align}
    \hxx_1^{(\omega)}(\xx_t) - \hxx_0^{(\omega)}(\xx_t) = \omega(\hxx_1(\xx_t) - \hxx_0(\xx_t)).
    \label{eq:cfg}
\end{align}
%

\noindent
\textbf{Final denoising step.} We found that applying a single denoising step to the solution of SB ODE effectively reduces residual noise, resulting in improved translation quality. To sum up, we describe the pseudo-code for the proposed method in Algorithm~\ref{alg:fasti2i}.

\begin{algorithm}[t]
    \centering
    \caption{Decomposed Schr\"{o}dinger Bridge ODE}
    \label{alg:fasti2i}
    \begin{algorithmic}[1]
    \Require {Encoder $\mathcal{E}$, Decoder $\mathcal{D}$, Source image $\xx_0$, text embeddings $c_0, c_1$, text-conditional VP diffusion noise predictor $\bm{\epsilon}_\theta$, CFG scale $\omega$, initial SB time $t_0$, SB variance $\tau$, NFE budget $N$.}
    \State $\hxx_0 \gets \mathcal{E}(\hxx_0)$
    \State $\hep_{\VP}^0(\cdot) \gets \bm{\epsilon}_\theta(\cdot,c_0)$, $\hep_{\VP}^1(\cdot) \gets \bm{\epsilon}_\theta(\cdot,c_1)$
    \State $\xx_{t_0} \gets (1-t_0)\xx_0 + \sqrt{t_0(1-t_0)\tau}\bm{\epsilon}$, \ $\bm{\epsilon}\sim\mathcal{N}(\bm{0},\bm{I})$
    \For{$i=1$ {\bfseries to} $N$}
        \State $t_i \gets t_0 + (1-t_0)(i-1)/N$
        \State $\sigma_{t_i} \gets (t_i(1-t_i)\tau)^{1/2}$
        \State $\abar_{s_i} \gets 1/(\sigma_{t_i}^2+1)$
        \State $\yy_{s_i} \gets \xx_{t_i} / \sqrt{\abar_{s_i}}$
        \State $\hxx_0^{(\omega)}(\xx_{t_i}) \gets \omega (\yy_{s_i} - \sqrt{1 - \abar_{s_i}} \hep^0_{\VP}(\yy_{s_i})) / \sqrt{\abar_{s_i}}$
        \State $\hxx_1^{(\omega)}(\xx_{t_i}) \gets \omega (\yy_{s_i} - \sqrt{1 - \abar_{s_i}} \hep^1_{\VP}(\yy_{s_i})) / \sqrt{\abar_{s_i}}$
        \State $\hep(\xx_{t_i}) \gets \hep_{\VP}^0(\yy_{s_i})$ \textbf{if} $t_i < 0.5$ \textbf{else} $\hep_{\VP}^1(\yy_{s_i})$
        \State $\vv(\xx_{t_i},t_i) \gets \frac{(0.5-t)\sqrt{\tau}}{\sqrt{t(1-t)}} \hep(\xx_t) + \hxx_1^{(\omega)}(\xx_t) - \hxx_0^{(\omega)}(\xx_t)$
        \State $\xx_{t_{i+1}} \gets \xx_{t_i} + \vv(\xx_{t_i}, t_i) / N$
    \EndFor
    \State {\bfseries return} $\mathcal{D}(\xx_1)$
    \end{algorithmic}
\end{algorithm}

\begin{table*}[t]
\centering
\resizebox{\linewidth}{!}{
\begin{tabular}{cccccgccccgccccg}
\toprule
 & \multicolumn{5}{c}{\textbf{Cat $\rightarrow$ Dog}} & \multicolumn{5}{c}{\textbf{Horse $\rightarrow$ Zebra}} & \multicolumn{5}{c}{\textbf{Dog $\rightarrow$ Wild*}} \\
\cmidrule(l){2-6} \cmidrule(l){7-11} \cmidrule(l){12-16}
 & Source & DDIB & PnP & SDEdit & Ours & Source & DDIB & PnP & SDEdit & Ours & Source & DDIB & PnP & SDEdit & Ours \\
\midrule
FID $\downarrow$
& 189.7 & 135.6 & 167.1 & \underline{123.2} & \textbf{113.2} & 258.8 & \underline{104.5} & 124.4 & 160.0 & \textbf{96.18} & 177.0 & \underline{114.4} & 239.8 & 156.3 & \textbf{94.79}   \\
KD $\downarrow$
& 3.296 & 5.993 & \textbf{4.109} & 5.187 & \underline{4.297} & 14.66 & \underline{2.505} & 2.798 & 3.454 & \textbf{1.910} & 4.909 & 6.350 & 5.934 & \underline{5.505} & \textbf{4.841} \\
DINOv2 FD$\downarrow$
& 3001.0 & 2329.7 & 2665.5 & \underline{2068.8} & \textbf{1813.3} & 2945.5 & \underline{902.5} & 999.4 & 1090.0 & \textbf{764.1} & 3299.8 & \underline{2973.4} & 2991.4 & 3354.3 & \textbf{2151.8} \\
LPIPS $\downarrow$
& - & 0.444 & \textbf{0.357} & 0.434 & \underline{0.418} & - & 0.479 & \textbf{0.284} & 0.448 & \underline{0.432} & - & 0.434 & \textbf{0.354} & 0.424 & \underline{0.396} \\
Structure Dist. $\downarrow$
& - & 0.081 & \textbf{0.019} & \underline{0.077} & \underline{0.077} & - & 0.184 & \textbf{0.022} & \underline{0.180} & 0.175 & - & 0.163 & \textbf{0.018} & 0.169 & \underline{0.161} \\
CLIP Score $\uparrow$
& 24.30 & 24.71 & 24.58 & \textbf{24.79} & \underline{24.78} & 22.33 & 22.94 & 22.19 & \underline{23.08} & \textbf{23.14} & 20.52 & \textbf{20.77} & \underline{20.64} & 20.42 & 20.62 \\
ImageReward $\uparrow$
& -2.147 & \underline{0.097} & -1.359 & -0.020 & \textbf{0.186} & -2.231 & -0.354 & \underline{-0.343} & -0.621 & \textbf{-0.280} & -0.509 & -0.316 & \underline{-0.264} & -0.642 & \textbf{-0.167}\\
\bottomrule
\end{tabular}
}
\caption{Quantitative Comparison for three image-to-image translation tasks (8 NFEs). \textbf{Bold} represents the best, and \underline{underline} denotes the second best method. * When we compute CLIP Score and ImageReward for wild domain, we use ``a photo of a wild feline'' as a prompt.}
\label{tab:eval}
\vspace{-2mm}
\end{table*}




\begin{table*}[ht]
\centering
\begin{minipage}{0.48\textwidth}
\includegraphics[width=\linewidth]{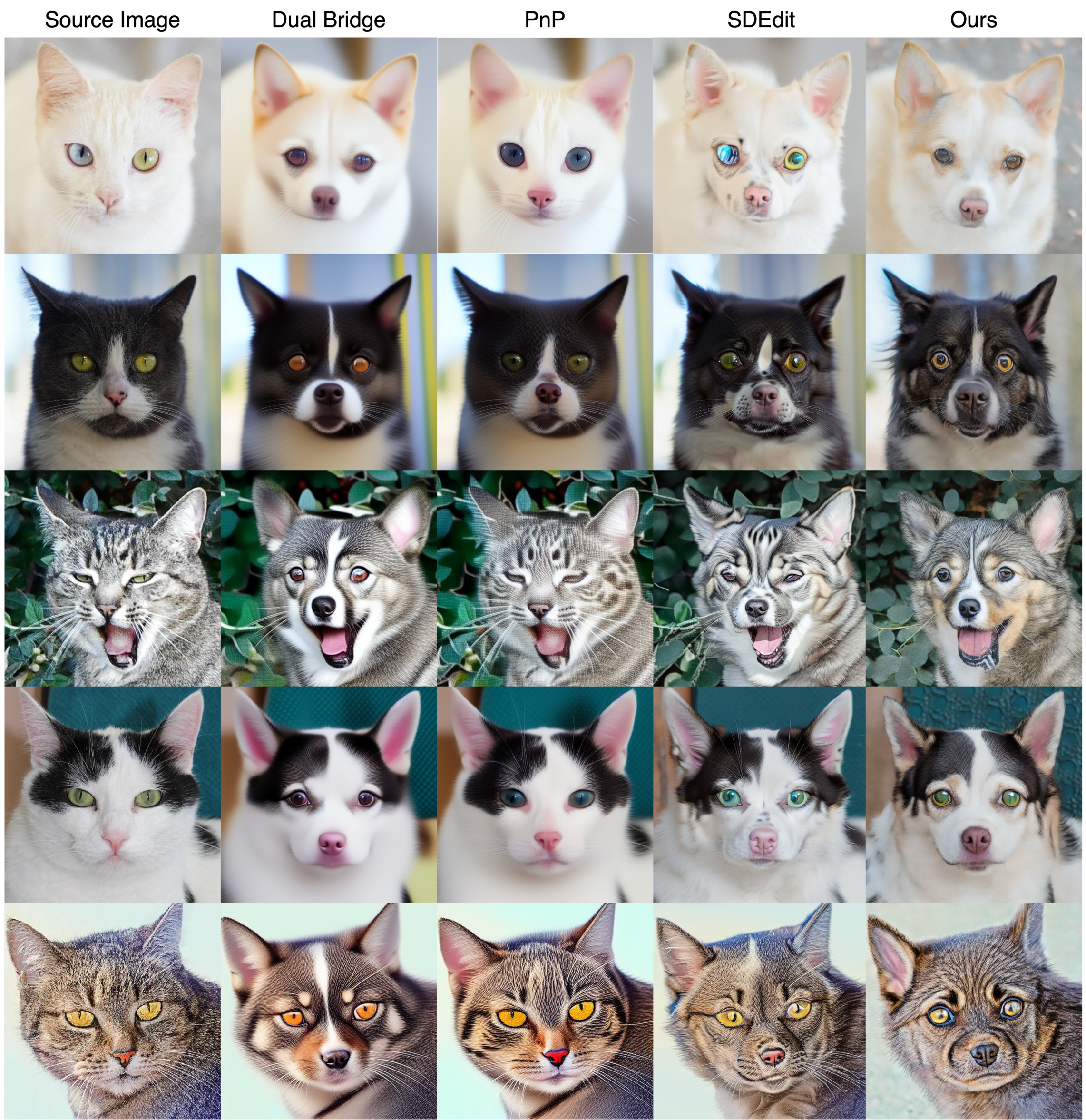}
\captionof{figure}{Qualitative comparison for Cat2Dog (8 NFEs)}
\label{fig:cat2dog}
\end{minipage} \hfill
\begin{minipage}{0.48\textwidth}
\includegraphics[width=\linewidth]{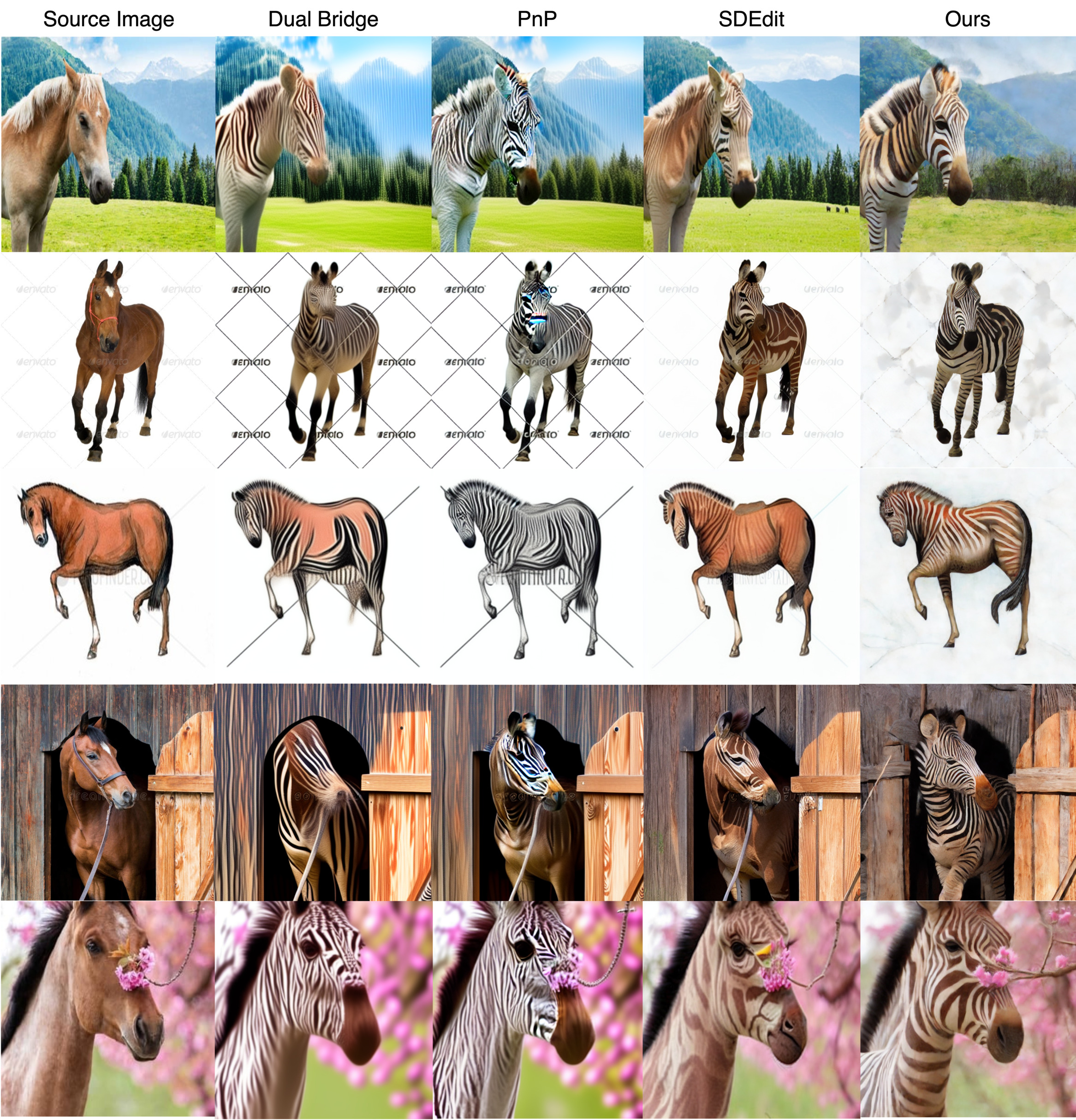}
\captionof{figure}{Qualitative comparison for Horse2Zebra (8 NFEs)}
\label{fig:horse2zebra}
\end{minipage}
\vspace{-2mm}
\end{table*}



\section{Experiments} \label{sec:experiments}

\subsection{Fast Unpaired I2I Translation}

\textbf{Baselines.}
We address the unpaired image-to-image translation problem using a pre-trained Stable Diffusion 1.5. As our method solely uses an ODE, and does not rely on feature or attention manipulation, we set Dual Bridge \citep{su2023ddib} and SDEdit \citep{meng2022sdedit} using SD 1.5 as comparable baselines.
Also, we compare with PnP \citep{tumanyan2023pnp} which could be interpreted as Dual Bridge with diffusion feature and self-attention injection.
To ensure a fair comparison, we employ the same text embeddings, numer of neural function evaluation (NFE), and CFG scale across all baselines. Specifically, for Dual Bridge and PnP, we allocate half of the NFEs to each of the inversion and sampling processes. For SDEdit, we use an initial SNR that matches the initial SNR of the proposed method. For more details, please refer to Appendix. In the main experiment, we use 8 NFEs for all cases, unless specified otherwise.

\noindent
\textbf{Datasets.}
In this paper, we examine the proposed method for Cat $\rightarrow$ Dog, Horse $\rightarrow$ Zebra, and Dog $\rightarrow$ Wild tasks with $512 \times 512$ resolution images. For each task, we leverage AFHQ and images sampled from LAION-5B dataset for both text prompt optimization and evaluation. Specifically, we select 1k images from training set for the text prompt optimization, and uses all images from validation set for the image-to-image translation. For more details, please refer to Appendix.

\noindent
\textbf{FID vs. NFEs.}
To demonstrate that our method provides a better trade-off between translation speed and translated sample quality, in Figure~\ref{fig:nfe}, we plot FID as a function of NFEs per sample. Indeed, LSB ODE outperforms DDIB and SDEdit by a large margin in the small NFE regime (NFE $\leq 10$). Furthermore, translation quality is improved or maintained with even more NFEs.


\noindent
\textbf{Quantitative comparison.}
To evaluate the quality of image translation, we leverage statistical distances between target images and translated images called FID \citep{heusel2017fid}, KD \citep{binkowski2018demystifying}, and DINOv2 FD \citep{stein2023exposing}. For the shape and background consistency between source images and translated images, we use DINO structure distance \citep{tumanyan2022splicing} and LPIPS \citep{zhang2018lpips}. Finally, to demonstrate whether indented semantic is well reflected after translation, we evaluate CLIP score \citep{radford2021learning} and ImageReward \citep{xu2024imagereward}. 
As shown in Table~\ref{tab:eval}, proposed method shows the highest fidelity and the best alignment with targeted semantic, while maintaining structural information and background better than baselines.
Meanwhile, PnP achieves the lowest Structure Distance and LPIPS scores, but this is primarily because PnP produces outputs nearly identical to the source images, resulting in minimal or no changes.

\noindent
\textbf{Qualitative comparison.}
We visualize the translated images by baselines method in Figure~\ref{fig:cat2dog} and ~\ref{fig:horse2zebra}, which is well aligned with the quantitative evaluation.
Specifically, SDEdit tends to preserve structural elements from the source image due to the lack of a repelling force from the source domain. Although this issue could be mitigated by reducing the initial SNR, doing so would increase the number of function evaluations (NFEs) required to sample from the data distribution.
On the other hand, Dual Bridge produces saturated images due to error accumulation when solving the PF-ODE with low NFEs. Similarly, PnP encounters the same type of error as Dual Bridge, leading to suboptimal results.

\begin{figure}
    \centering
    \includegraphics[width=\linewidth]{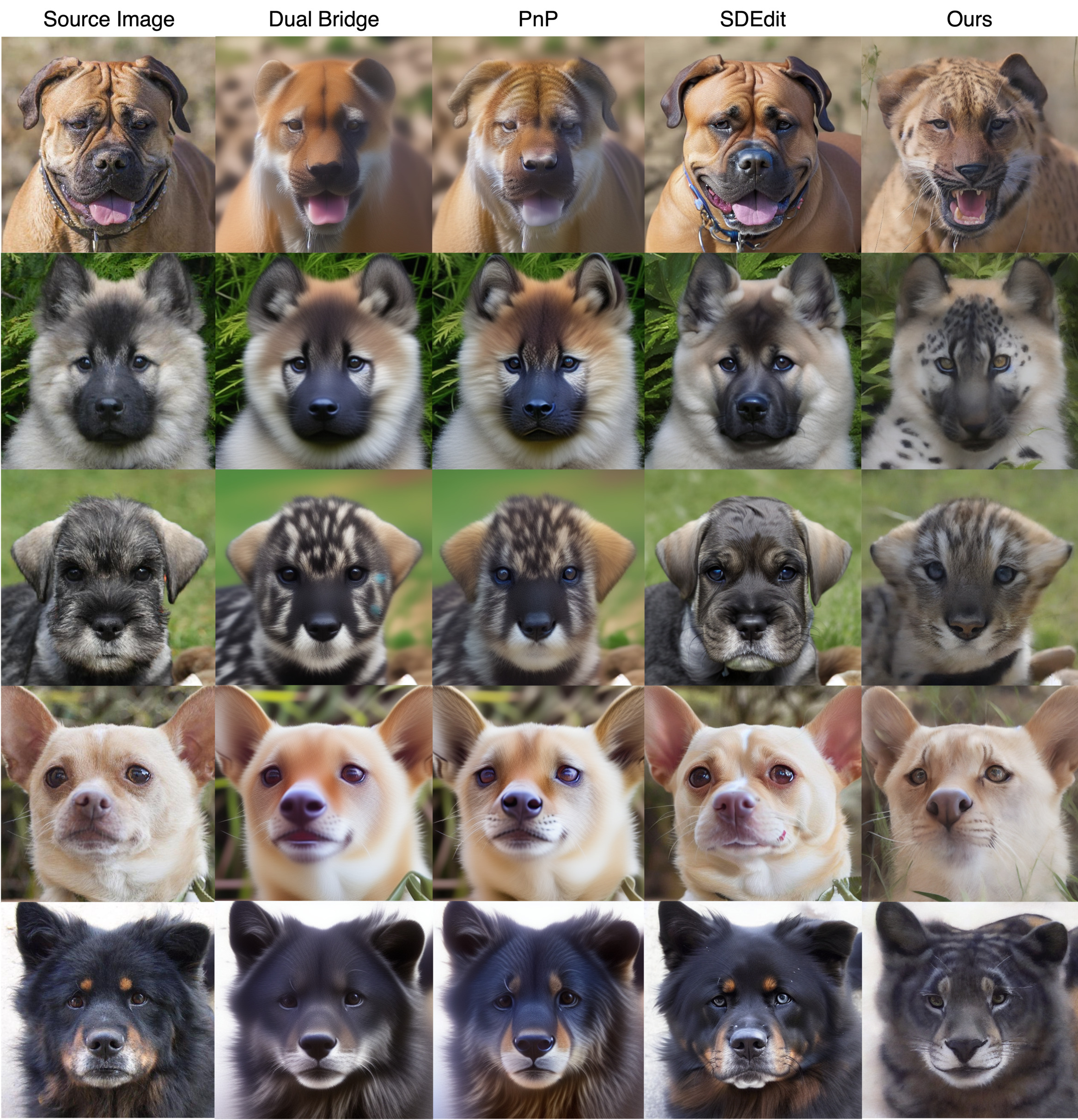}
    \caption{Qualitative comparison for Dog2Wild (8 NFEs)}
    \label{fig:dog2wild}
\end{figure}

\subsection{Ablation Study}

\begin{figure}
    \includegraphics[width=0.95\linewidth]{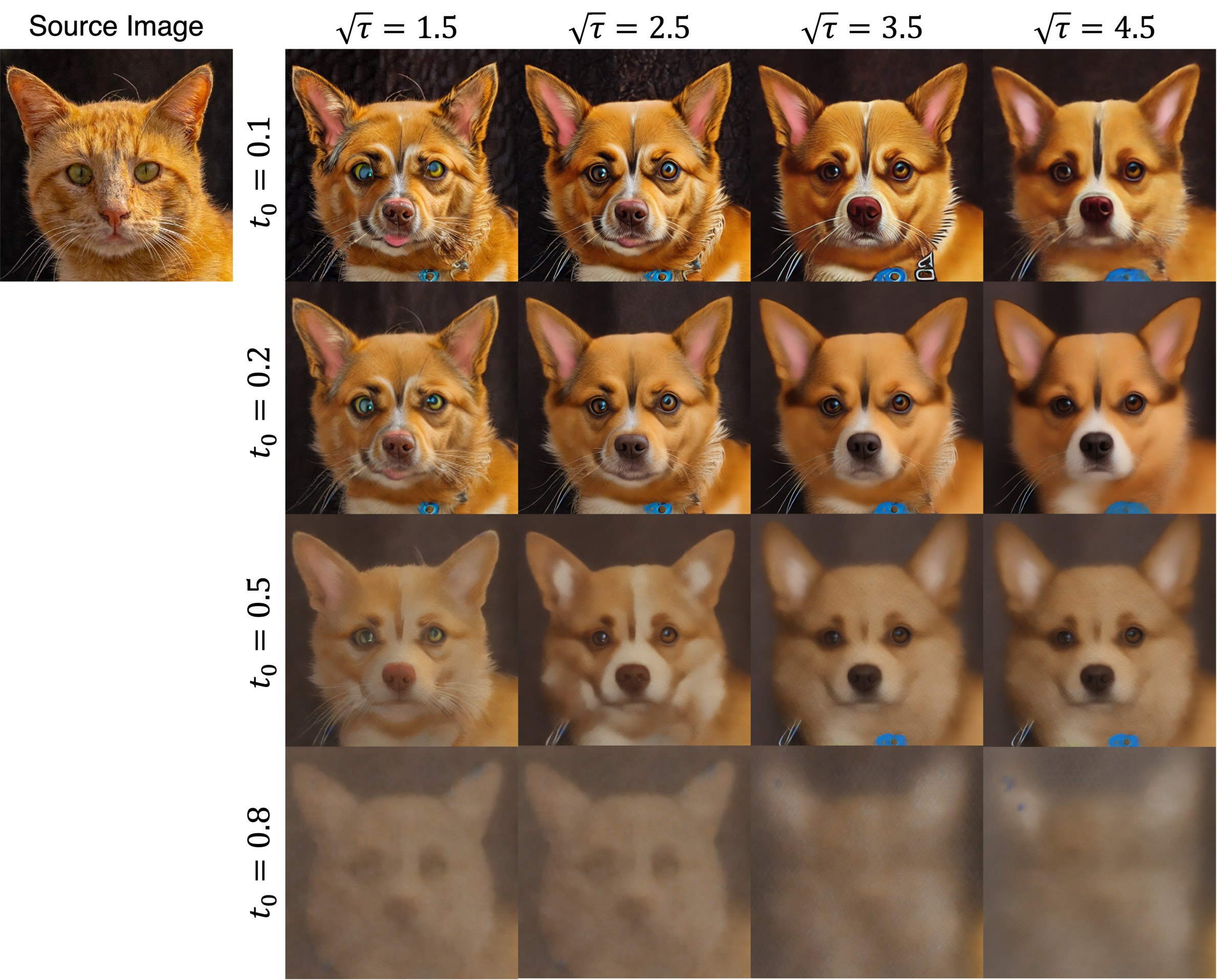} \hfill
    \caption{Solutions with varying $\tau$ and $t_0$.}
    \label{fig:t0_tau}
    \vspace{-2mm}
\end{figure}

\noindent
\textbf{Effect of $\tau$ and initial time $t_0$.}
To demonstrate the effects of $\tau$ and initial time $t_0$ on the LSB, we conduct an ablation study on the Cat$\rightarrow$Dog task, with the results visualized in Figure~\ref{fig:t0_tau}.
As $t_0$ increases to 1, the resulting images become progressively blurrier and lose details from the source image (see last row). This is because we initialize $\xx_{t_0}$ according to \eqref{eq:init}, and a larger $t_0$ results in a significant SNR mismatch between $\xx_t$ and the training samples input to the VP diffusion model.
Furthermore, increasing $\tau$ introduces additional noise into the the translation trajectory, resulting in blurrier outputs due to a blurred posterior mean computed at the final denoising step. 
Conversely, decreasing $\tau$ introduces insufficient noise for effective translation, leading to outputs that retain too many details from the source image and disrupt the intended image translation.
Based on the findings from this ablation study, we set $t_0 = 0.2$ and $\sqrt\tau = 2.5$ with 8 NFEs for all tasks in the main experiment.

\noindent
\textbf{LSB component ablation.}
To assess the impact of each component designed for practical considerations (see Sec~\ref{sec:practical}), we sequentially remove the time-dependent noise predictor, CFG, the final denoising step, and SNR matching when performing horse $\rightarrow$ zebra translation task. All other hyperparameters including NFE, $t_0$ and $\tau$ are kept consistent with those in the main experiment.
As shown in Table~\ref{tab:ablation}, each component is essential for enhancing the quality of image translation. Notably, SNR matching makes significant improvement in achieving solutions that are closely aligned with the target distribution.
\add{Figure~\ref{fig:abl_snr} in Appendix provides a qualitative comparison illustrating the effects of applying SNR matching.} Without SNR matching, the predicted VP score function lacks accuracy, leading the solution of the ODE to deviate from the target distribution as a result.

\begin{table}[t]
\centering
\resizebox{\linewidth}{!}{
\begin{tabular}{ccccc}
\toprule
\textbf{SNR Matching} & \textbf{Time-dependent} $\bm{\epsilon}(\cdot)$ & \textbf{CFG} & \textbf{Denoising Step} & \textbf{FID $\downarrow$} \\
\midrule
\cmark & \cmark & \cmark & \cmark & \textbf{94.79}\\
\xmark & \cmark & \cmark & \cmark & 284.7\\
\xmark & \xmark & \cmark & \cmark & 285.5\\
\xmark & \xmark & \xmark & \cmark & 288.7\\
\xmark & \xmark & \xmark & \xmark & 413.1\\
\bottomrule
\end{tabular}}
\caption{Ablation study for components induced by practical considerations. FID is evaluated for Horse $\rightarrow$ Zebra task.}
\label{tab:ablation}
\end{table}

\section{Conclusion} \label{sec:conclusion}

In this paper, we proposed Latent Schr\"{o}dinger Bridge (LSB) ordinary differential equation formulation for fast unpaired image-to-image translation via pretrained latent diffusion models. We decomposed the SB probability flow ODE velocity into a linear combination of three terms -- source domain image predictor, target domain image predictor, and Gaussian noise predictor. We then developed change-of-variables formulae for mitigating signal-to-noise ratio mismatch between diffusion and SB variables. Finally, we proposed various practical techniques for making our method work using a pretrained Stable Diffusion. We verifed that our method beats diffusion-based I2I methods such as SDEdit or Dual Diffusion Implicit Bridge with significantly fewer neural net evaluations.


\clearpage

{
    \small
    \bibliographystyle{ieeenat_fullname}
    \bibliography{main}
}

\newpage

\appendix

\clearpage
\setcounter{page}{1}
\maketitlesupplementary

\section{Generation with Optimized Text embedding}
\begin{figure}[t]
    \centering
    \includegraphics[width=\linewidth]{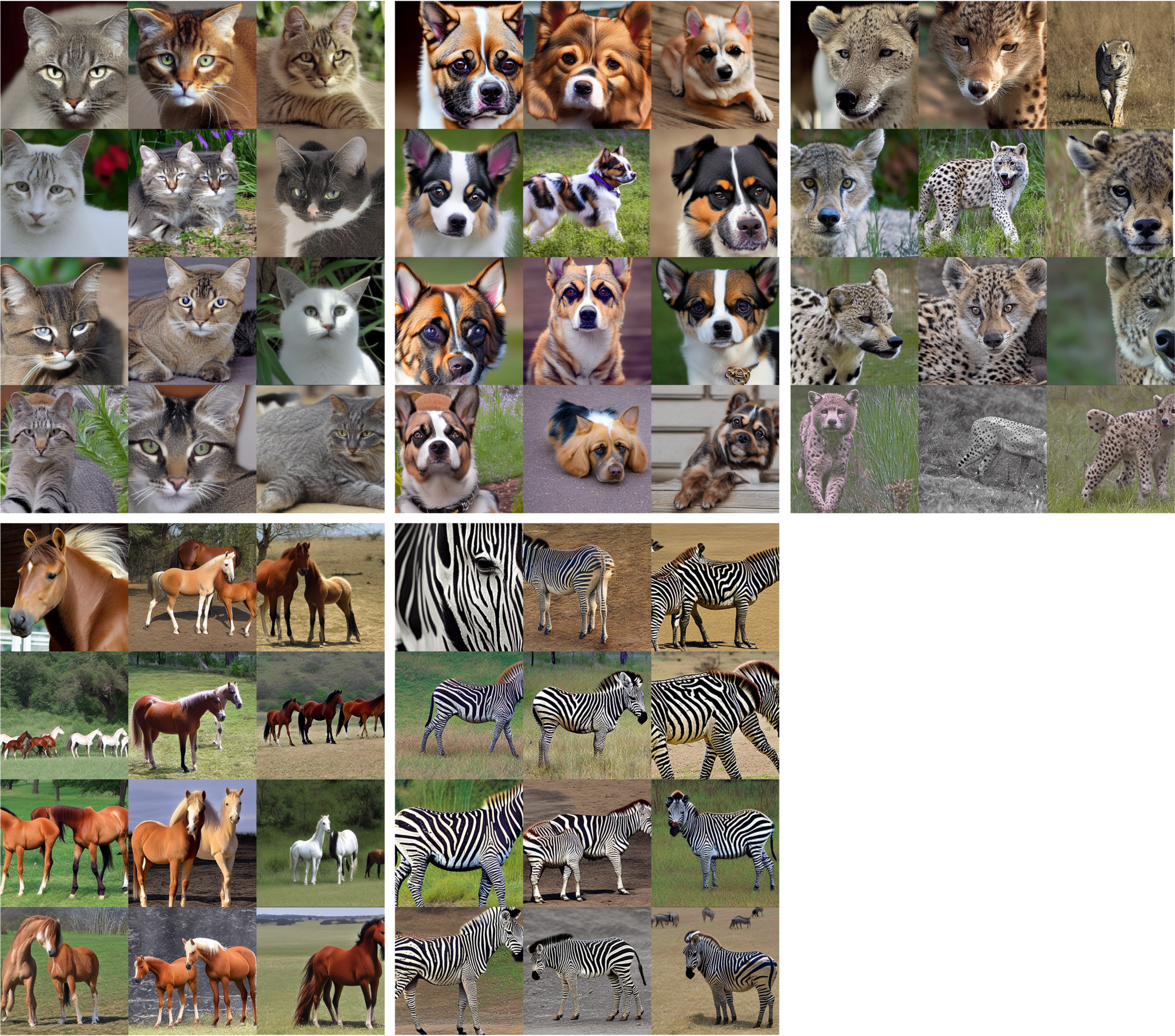}
    \caption{Uncurated samples generated by VP diffusion with optimized text embedding.}
    \label{fig:textinv_sample}
\end{figure}

In this study, we investigate an unsupervised image-to-image translation task where specific dataset descriptions are unavailable. To address this, we leverage textual embedding to obtain a text embedding capable of guiding the pre-trained VP diffusion model to sample from a specified domain. Figure~\ref{fig:textinv_sample} demonstrates that the VP diffusion model, using optimized text embeddings, successfully replicates samples from each domain (clockwise from upper left: cat, dog, wild, zebra, horse) without any supervision on text descriptions.

\section{Omitted Proofs} \label{append:proofs}

\subsection{Proof of Proposition \ref{prop:gamma}}

\begin{proof}
    It is known that the conditional vector field
    \begin{align*}
        \textstyle \uu(\xx_t | \xx_0, \xx_1, t) \coloneqq \frac{(1/2-t)}{t(1-t)}(\xx_t - \bm{\mu}_t) + (\xx_1 - \xx_0),
    \end{align*}
    generates the probability path
    \begin{align*}
        d\PP_{t|01}(\xx_t | \xx_0,\xx_1) = \mathcal{N}(\xx_t | \bm{\mu}_t, \sigma_t^2 \bm{I}).
    \end{align*}
    For instance, see Sec. 3.2.4 of \citep{tong2023cfm}. By Thm. 3.1 in \citep{tong2023cfm}, the ODE with velocity \eqref{eq:gamma_decomp_vel} generates the probability path $\PP_t$ derived from \eqref{eq:gamma_joint_cond}. In particular, $\PP_t$ at $t = 0$ and $1$ are equal to data distributions $\PP_0$ and $\PP_1$, resp.
\end{proof}

\subsection{Proof of Proposition \ref{prop:sanity}}

\begin{proof}
    In the case $\tau = 0$, $\Gamma_{01} = \PP_0 \otimes \PP_1$, and $\PP_1$ is standard normal, $\vv$ is the solution to the regression problem
    \begin{align*}
        \min_{\vv} \EE_{(\xx_1,\xx_0) \sim \Gamma_{01}} [ \| (\xx_1-\xx_0) - \vv(\xx_t,t) \|_2^2]
    \end{align*}
    where $\xx_t = (1-t)\xx_0 + t\xx_1$. This is training objective for Rectified Flow \citep{liu2022rf}, and Rectified Flow ODE is equivalent to the diffusion probability flow ODE (see Section 2 of \citep{kim2024reflow}).
\end{proof}

\subsection{VP Diffusion and SB SNR Matching}

Observe that SB and VP diffusion variables are given by
\begin{align*}
    &\xx_t = \bm{\mu}_t + \sigma_t \bm{\epsilon}, \\
    &\yy_s = \sqrt{\abar_s} \xx_0 + \sqrt{1 - \abar_s} \bm{\epsilon}.
\end{align*}
If we equate the SNRs,
\begin{align}
    \SNR(\xx_t) = \sigma_t^{-2} = \abar_s / (1 - \abar_s) = \SNR(\yy_s)
\end{align}
we obtain
\begin{align}
\abar_s = 1/(\sigma_t^2+1).
\end{align}
In addition,
\begin{align}
    \xx_t &= \bm{\mu}_t + \sigma_t \bm{\epsilon} \\
    &= \bm{\mu}_t + (\sqrt{1 - \abar_s} / \sqrt{\abar_s}) \bm{\epsilon}
\end{align}
so we may set $\yy_s = \sqrt{\abar_s} \xx_t$ to match SB and VP forms.
 
\section{Implementation Details}
In this section, we provide implementation details for both text optimization and image to image translation. The codebase will be available on public \url{https://github.com/LatentSB/LatentSB}.

\subsection{VP diffusion model}
In this study, we utilize the pre-trained Stable Diffusion v1.5 model as the backbone for the VP  which serves as the predictor for the source image, target image, and Gaussian noise. Thus, we employ the pre-trained CLIP ViT/L-14 model as the text encoder. It is noteworthy that our framework is not restricted to a specific version of text-conditioned latent diffusion models; it could be extended to SDXL or SD3. However, we leave this exploration for future application.

\subsection{Text embedding optimization}
For the text embedding optimization, we follows experimental setup of the textual inversion~\cite{gal2022textual}. Specifically, in all tasks, we optimize the text embeddings of the text encoder (CLIP) using the AdamW with learning rate $1\times 10^{-4}$ and weight decay $1\times 10^{-2}$. The bath size is set to 1, and we train the embeddings for 1k iteration using 1k training samples. 
Although training converges quickly, additional iterations may be necessary if the given images for each domain are underrepresented in the pre-trained diffusion prior.

For the trainable text embedding, we use placeholder text ``!'' to represent the source domain and ``*'' for the target domain. Specifically, we utilize the template provided by the textual inversion example on HuggingFace, such as ``a photo of a *''. The trainable text embeddings for both domain are initialized with ``a photo''.

The training objective is identical to the denoising score matching used in the pre-training of the VP diffusion, defined as
\begin{align}
    \min_{V} \mathbb{E}_{\bm{\epsilon}\sim\mathcal{N}, t\sim U[0, T]} \| \bm{\epsilon} - \bm{\epsilon}_\theta(\xx_t, t, V) \|^2
\end{align}
where $V$ denotes trainable text embedding, $\mathcal{N}$ represents a normal distribution, and $U[0, T]$ denotes a uniform distribution over the diffusion time interval from 0 to $T$.

\subsection{Dataset}

\noindent
\textbf{Cat $\rightarrow$ Dog.} For the text optimization, we select 1k images from training set of AFHQ for both cat and dog domains. For the image translation, we use all images in validation set of AFHQ Cat. For the evaluation, we use all images in validation set of AFHQ Dog as target distribution.

\noindent
\textbf{Horse $\rightarrow$ Zebra.} For the text optimization, we use 1k images from training set of horse2zebra dataset from CycleGAN \citep{zhu2017cyclegan} for both horse and zebra domain. For the image translation, we use 209 images sampled from LAION-5B dataset with query ``horse''. For the evaluation, we use 173 images sampled from LAION-5B dataset with query ``zebra''.

\noindent
\textbf{Dog $\rightarrow$ Wild.} For the text optimization, we utilize 1k images from training set of AFHQ for both dog and wild domains. For the image translation and evaluation, we use all images in validation set of AFHQ Cat and Wild. Specifically, for the wild domain, we use a text prompt ``a photo of a wild feline'' when we evaluate CLIP Score and ImageReward.

\subsection{Metric}

\noindent
\textbf{FID \citep{heusel2017fid}} We compute Frechet Inception Distance (FID) using evaluation code from \url{https://github.com/ML-GSAI/EGSDE}. We set batch size to 50 and feature dimension to 2048. The activations are computed by InceptionV3. We use default parameters for all others.

\noindent
\textbf{KD \citep{binkowski2018demystifying}} We compute Kernel Distance (KD) using evaluation code from \url{https://github.com/layer6ai-labs/dgm-eval}. We set batch size to 50 and use default parameters for all others.

\noindent
\textbf{DINOv2 FD \citep{stein2023exposing}} We also compute FID score by leveraging DINOv2 as encoder which provides more richer evaluation. We use evaluation code from \url{https://github.com/layer6ai-labs/dgm-eval}. We use the default parameters.

\noindent
\textbf{LPIPS \citep{zhang2018lpips}} We use the official pytorch implementation from \url{https://github.com/richzhang/PerceptualSimilarity}. For the network architecture to extract activations, we use VGG network.

\noindent
\textbf{Structure Distance \citep{tumanyan2022splicing}} We compute the structure distance using loss function from \url{https://github.com/omerbt/Splice}. 

\noindent
\textbf{CLIP Score \citep{radford2021learning}} We compute the CLIP Score using implementation of \url{https://github.com/Lightning-AI/torchmetrics}. For the CLIP model, we utilize the ViT-base-patch16 variant and resize the input images to 224×224×3 before feeding them into the CLIP image encoder.

\noindent
\textbf{ImageReward \citep{xu2024imagereward}} We compute the ImageReward score using the official pytorch implementation from \url{https://github.com/THUDM/ImageReward}. 

\begin{figure*}[ht]
    \centering
    \includegraphics[width=\linewidth]{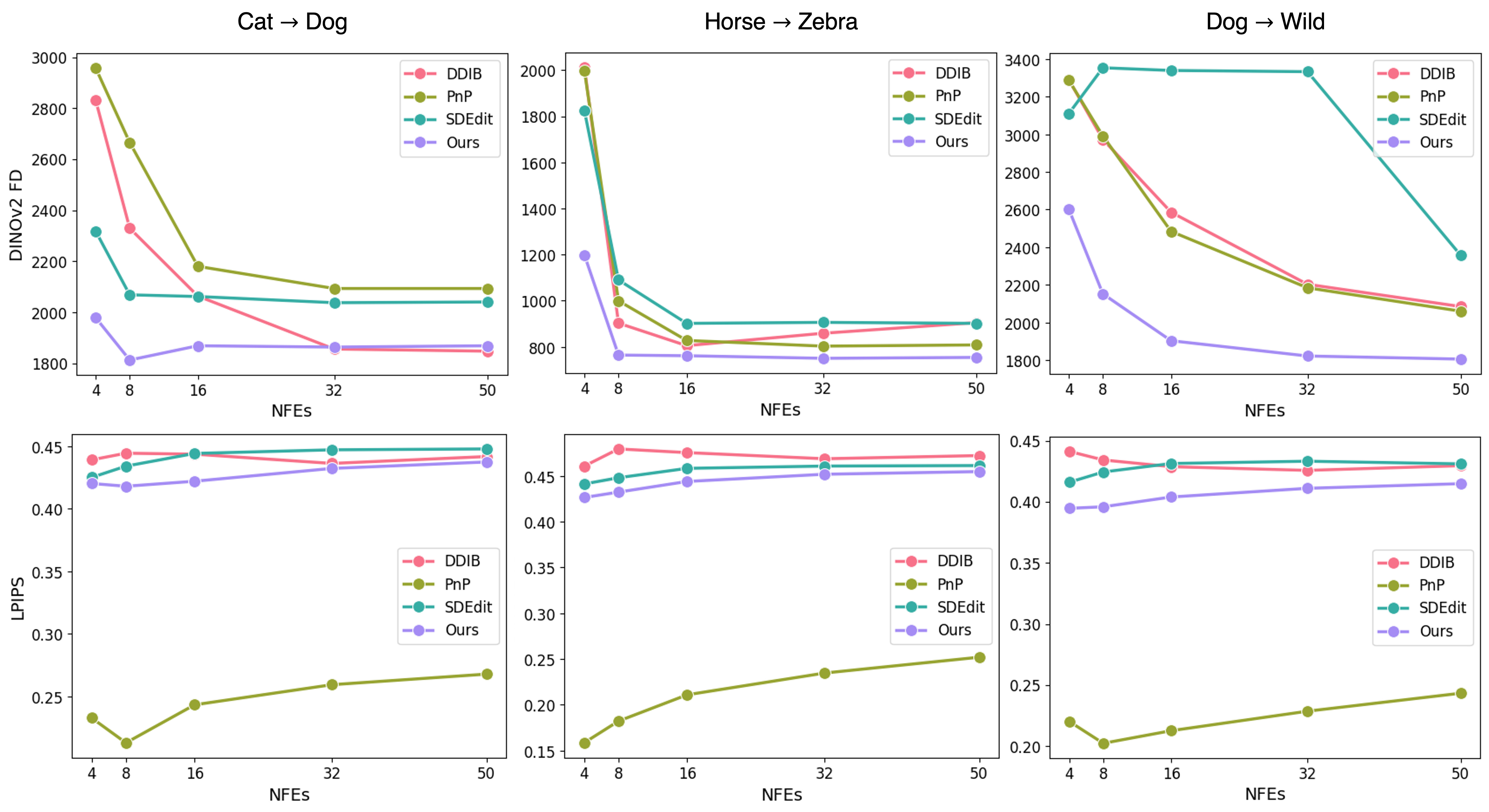}
    \caption{Quantitative comparison across NFEs: We evaluate FID, using DINOv2 as the encoder, and LPIPS on translated images from baseline methods across different NFEs. Results indicate that LSB ODE generates samples that more closely match the target distribution while preserving the shape and structure of the source images.}
    \label{fig:abl_others}
\end{figure*}

\subsection{Parameters for FID vs NFE comparison}
For the FID vs NFE comparison reported in Figure~\ref{fig:nfe}, we set the hyper-parameters such as CFG scale and $t_0$ as following, which are determined by grid search. For the proposed method, we allocate N-1 NFEs to solving the LSB ODE and use the remaining 1 NFE for the final denoising step, assuming a total of N NFEs for fair comparison. In contrast, for the Dual Bridge and PnP, we allocate N/2 NFEs for inversion and the remaining N/2 NFEs for the reverse sampling.

\begin{itemize}
    \item Task: Dog $\rightarrow$ Cat, Horse $\rightarrow$ Zebra
    \\DDIB (4-16\footnote{NFEs from 4 to 16.}): $\omega=5.0$
    \\DDIB (32, 50): $\omega=3.0$
    \\PnP (4-50): $\omega=11.0$
    \\SDEdit, Ours (4): $t_0=0.2$, $\sqrt\tau=1.5$, $\omega=11.0$
    \\SDEdit, Ours (8): $t_0=0.2$, $\sqrt\tau=2.5$, $\omega=11.0$
    \\SDEdit, Ours (16-50): $t_0=0.2$, $\sqrt\tau=3.5$, $\omega=11.0$
    \item Task: Dog $\rightarrow$ Wild
    \\DDIB (4-50): $\omega=5.0$
    \\PnP (4-50): $\omega=11.0$
    \\SDEdit, Ours (4): $t_0=0.2$, $\sqrt\tau=1.5$, $\omega=11.0$
    \\SDEdit, Ours (8): $t_0=0.2$, $\sqrt\tau=2.3$, $\omega=11.0$
    \\SDEdit, Ours (16-50): $t_0=0.2$, $\sqrt\tau=3.0$, $\omega=11.0$
\end{itemize}

\begin{figure}
    \centering
    \includegraphics[width=\linewidth]{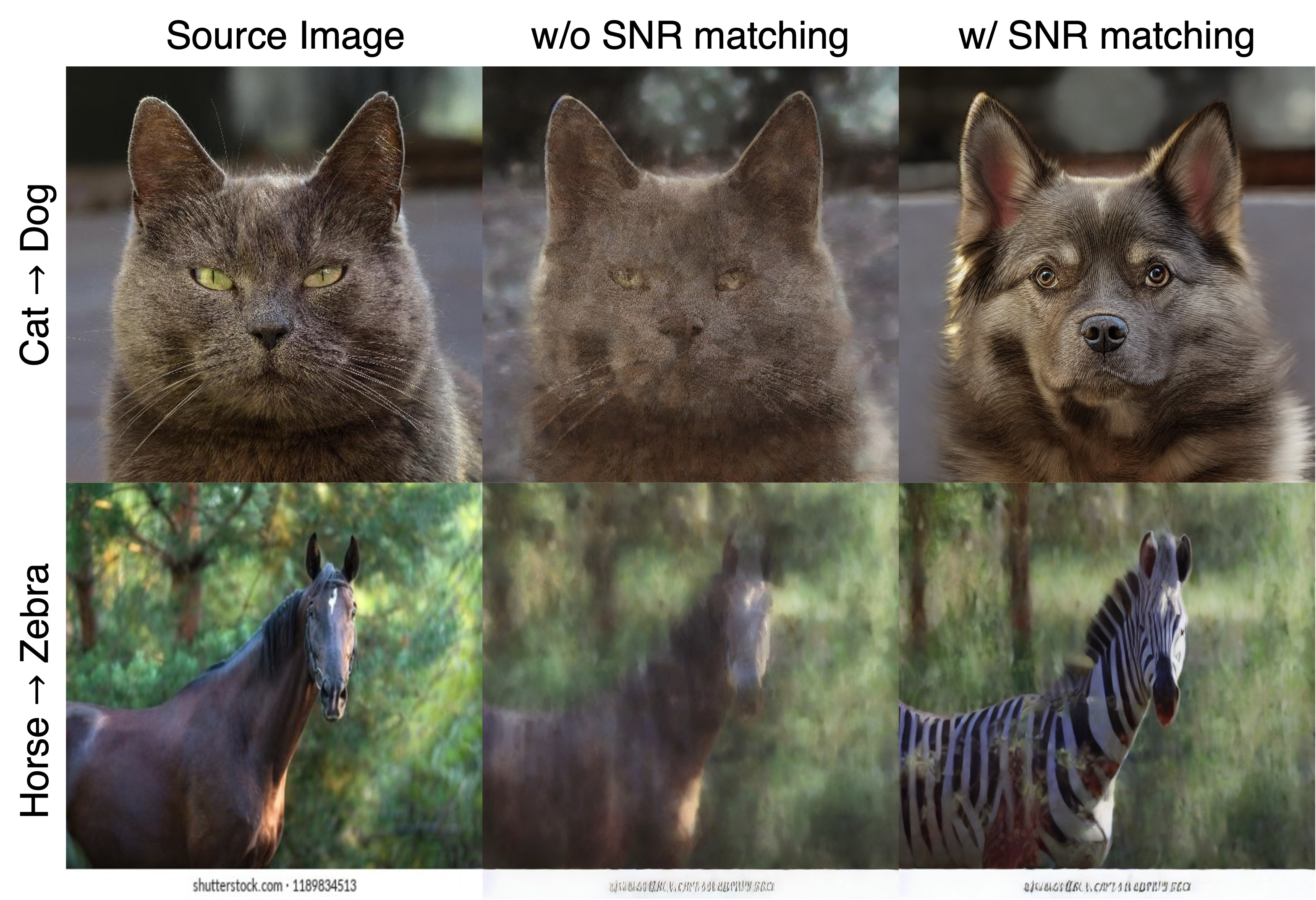}
    \caption{Translated image without and with SNR matching.}
    \label{fig:abl_snr}
    \vspace{-4.5mm}
\end{figure}

\begin{figure}
    \centering
    \includegraphics[width=\linewidth]{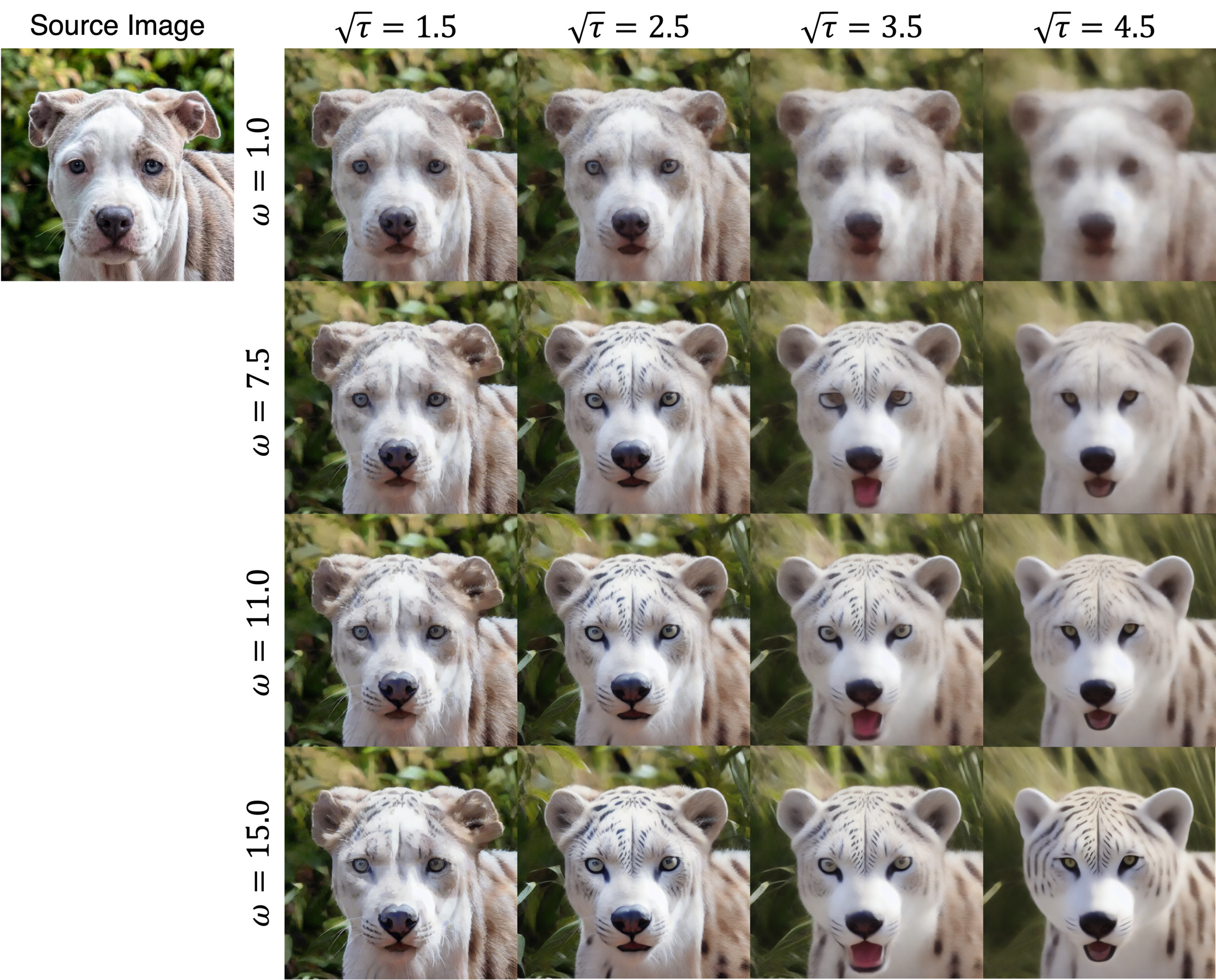}
    \caption{Solutions of LSB with varying $\omega$ and $\tau$.}
    \label{fig:abl_cfg_tau}
\end{figure}

\section{Ablation Study}
\subsection{Effect of SNR matching}
Figure~\ref{fig:abl_snr} illustrates translated images both without and with SNR matching. Without SNR matching, errors introduced by the pre-trained VP diffusion model cause samples to deviate significantly from the target distribution.

\subsection{Effect of CFG scale}
Figure~\ref{fig:abl_cfg_tau} provides solutions of LSB ODE with varying CFG scale and $\tau$. Without CFG scale (i.e. $\omega=1.0$), solutions tend not to capture target domain features. As $\omega$ increases, target domain features become more prominent. Notably, even with higher $\tau$ (indicating more noise in the sampling process), a larger $\omega$ helps reduce blur in the output. However, excessively high $\omega$ values can cause significant background alterations or introduce artifacts.

\section{Additional Quantitative Comparison}
In Figure~\ref{fig:abl_others}, we plot the DINOv2 FD and LPIPS across different NFEs for each baseline method. 
In terms of DINOv2 FD, LSB ODE outperforms DDIB and SDEdit for all cases. Specifically, LSB ODE shows a large margin compared to baselines in the small NFE region ($\leq 10)$ and shows maintained or improved performance with more NFEs, similar to FID. 
In contrast, for LPIPS calculated between source and translated images, PnP achieves the best score; however, as discussed in Section \ref{sec:experiments} and shown in Figures \ref{fig:cat2dog}–\ref{fig:dog2wild}, this result stems from minimal changes in the images due to feature and attention capture from inaccurate inversion with low NFEs. Excluding PnP, the LSB ODE yields lower LPIPS scores compared to DDIB and SDEdit, indicating that images translated by the LSB ODE retain perceptual similarity with the source images while effectively incorporating the target concept.

\section{Additional examples}
\begin{figure*}[t]
    \centering
    \includegraphics[width=\linewidth]{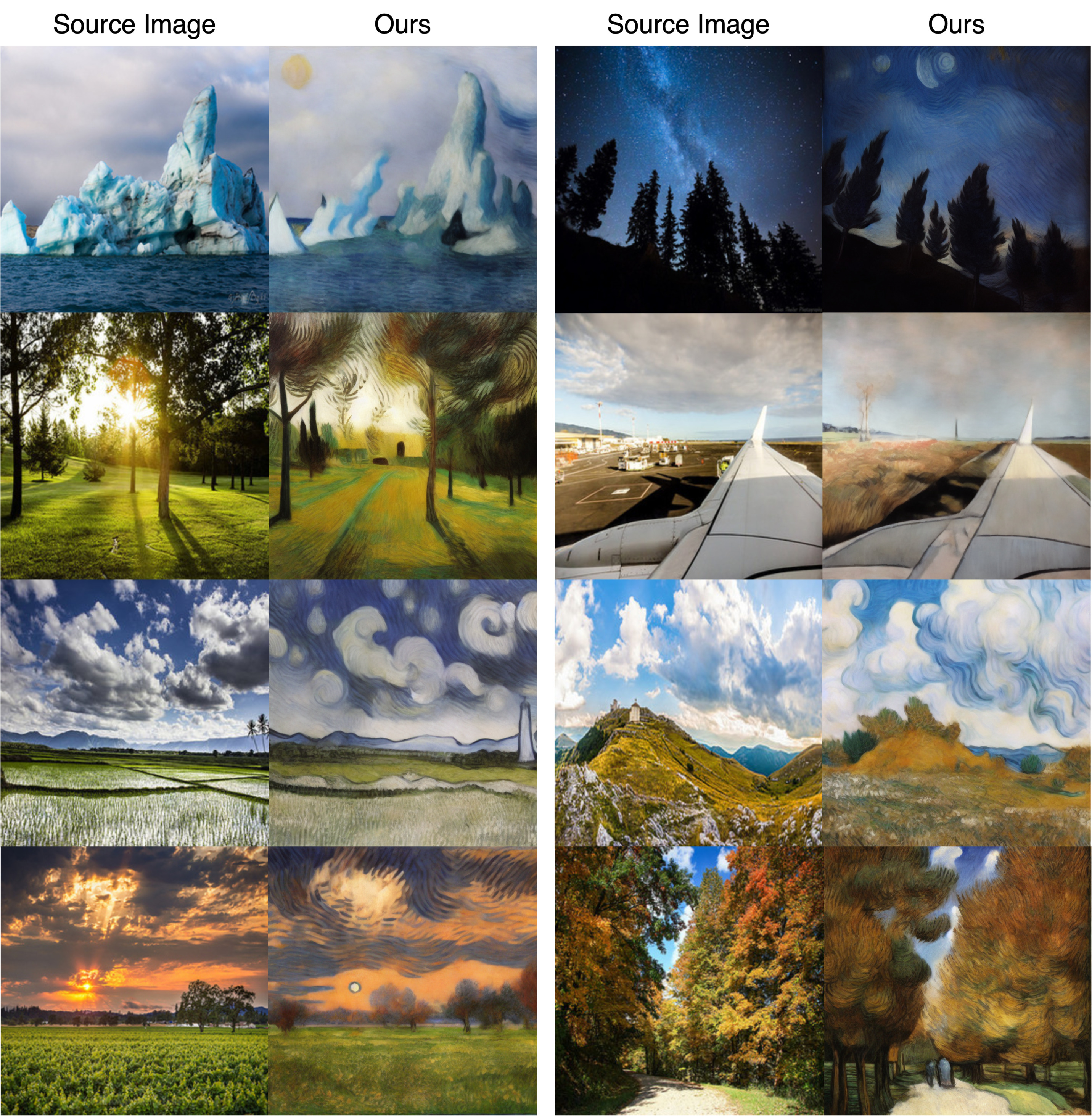}
    \caption{Image translation example by LSB for photo $\rightarrow$ VanGogh task (8 NFEs). We use cycleGAN vangogh2photo dataset.}
    \label{fig:photo2gogh}
\end{figure*}

\begin{figure*}[t]
    \centering
    \includegraphics[width=\linewidth]{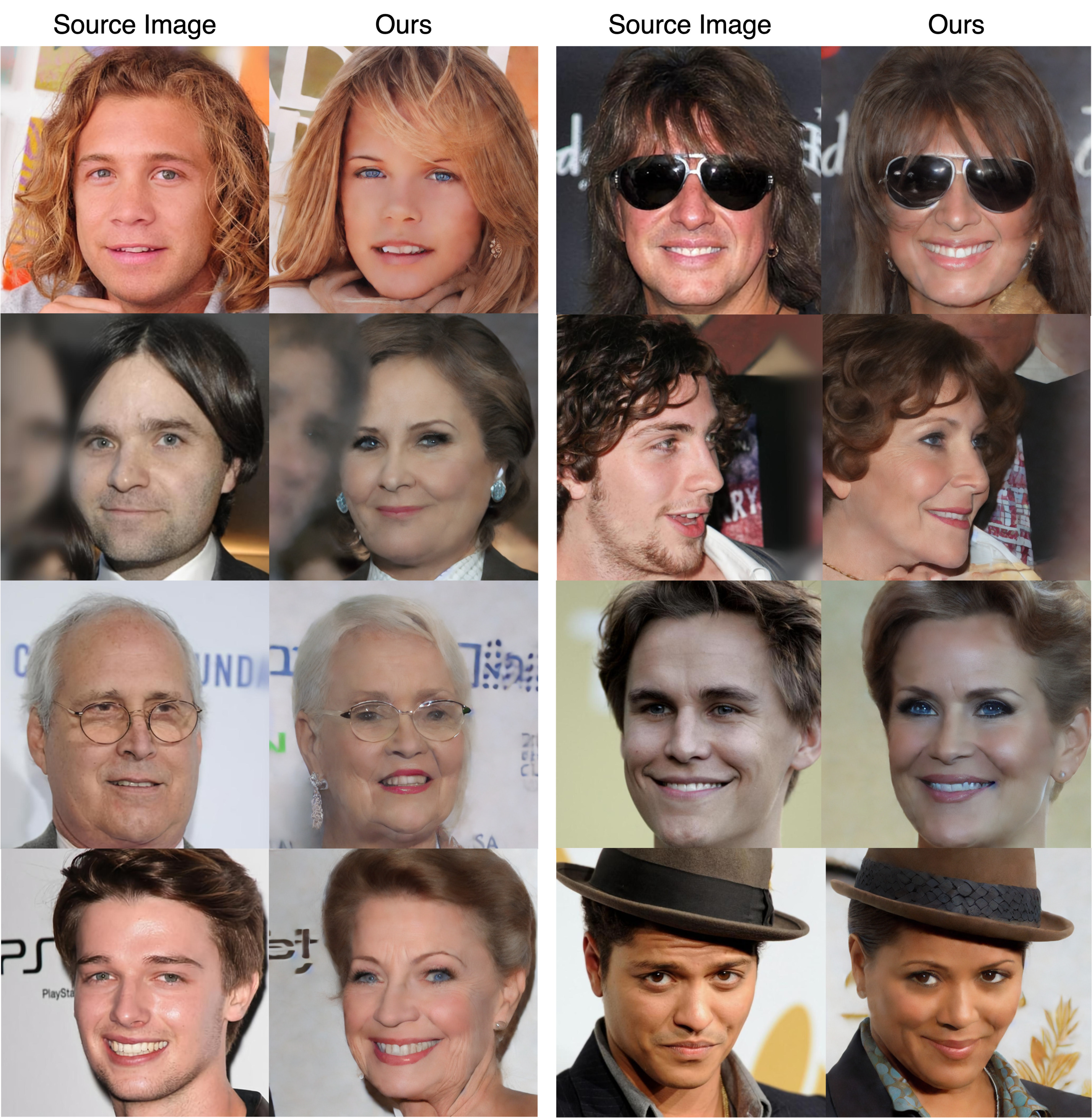}
    \caption{Image translation example by LSB for male $\rightarrow$ female task (8 NFEs). We use CelebA-HQ dataset.}
    \label{fig:male2female}
\end{figure*}

\begin{figure*}[t]
    \centering
    \includegraphics[width=\linewidth]{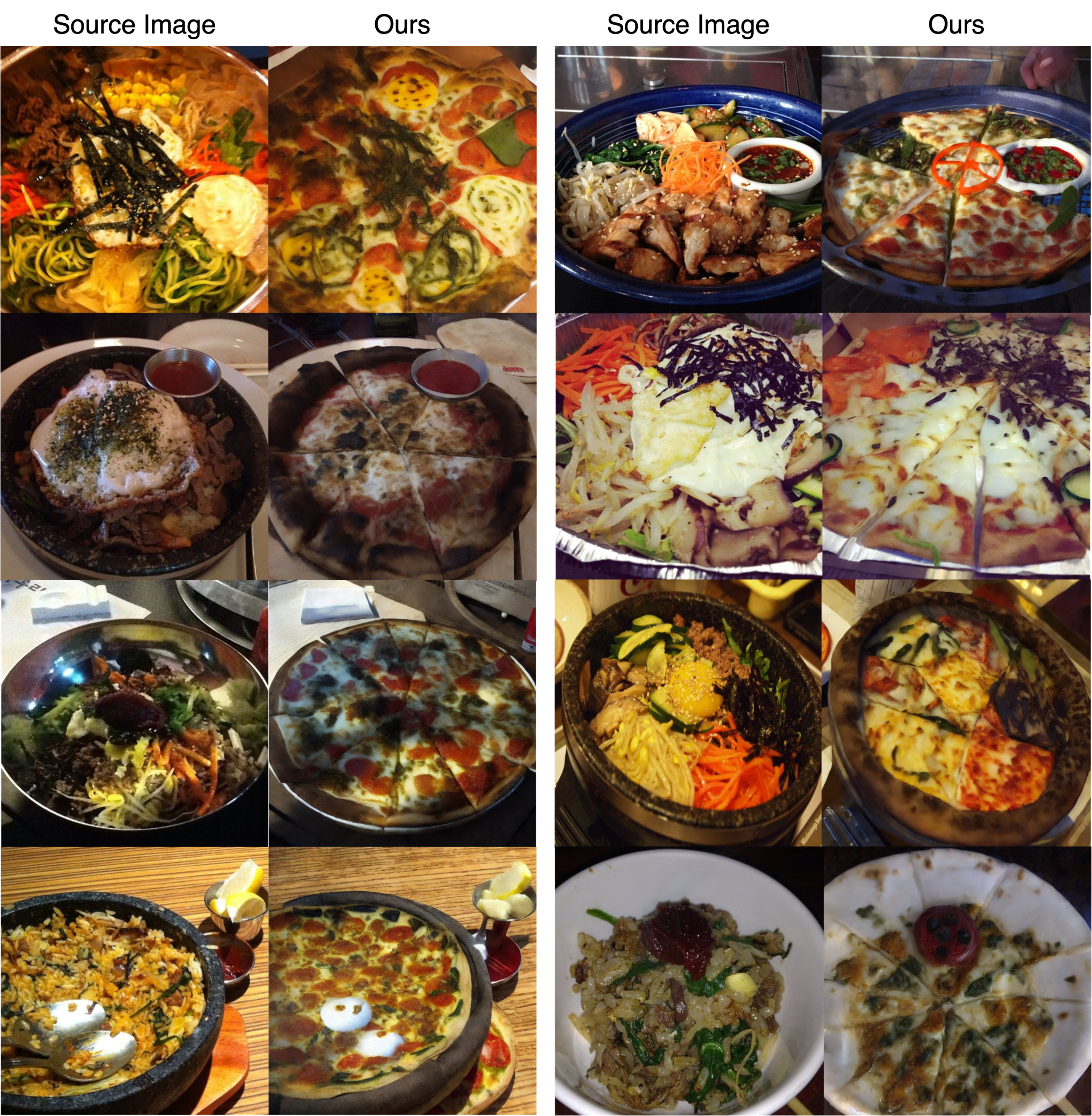}
    \caption{Image translation example by LSB for bibimbap $\rightarrow$ pizza task (8 NFEs). We use Food-101 dataset.}
    \label{fig:bibim2pizza}
\end{figure*}

\begin{figure*}[t]
    \centering
    \includegraphics[width=\linewidth]{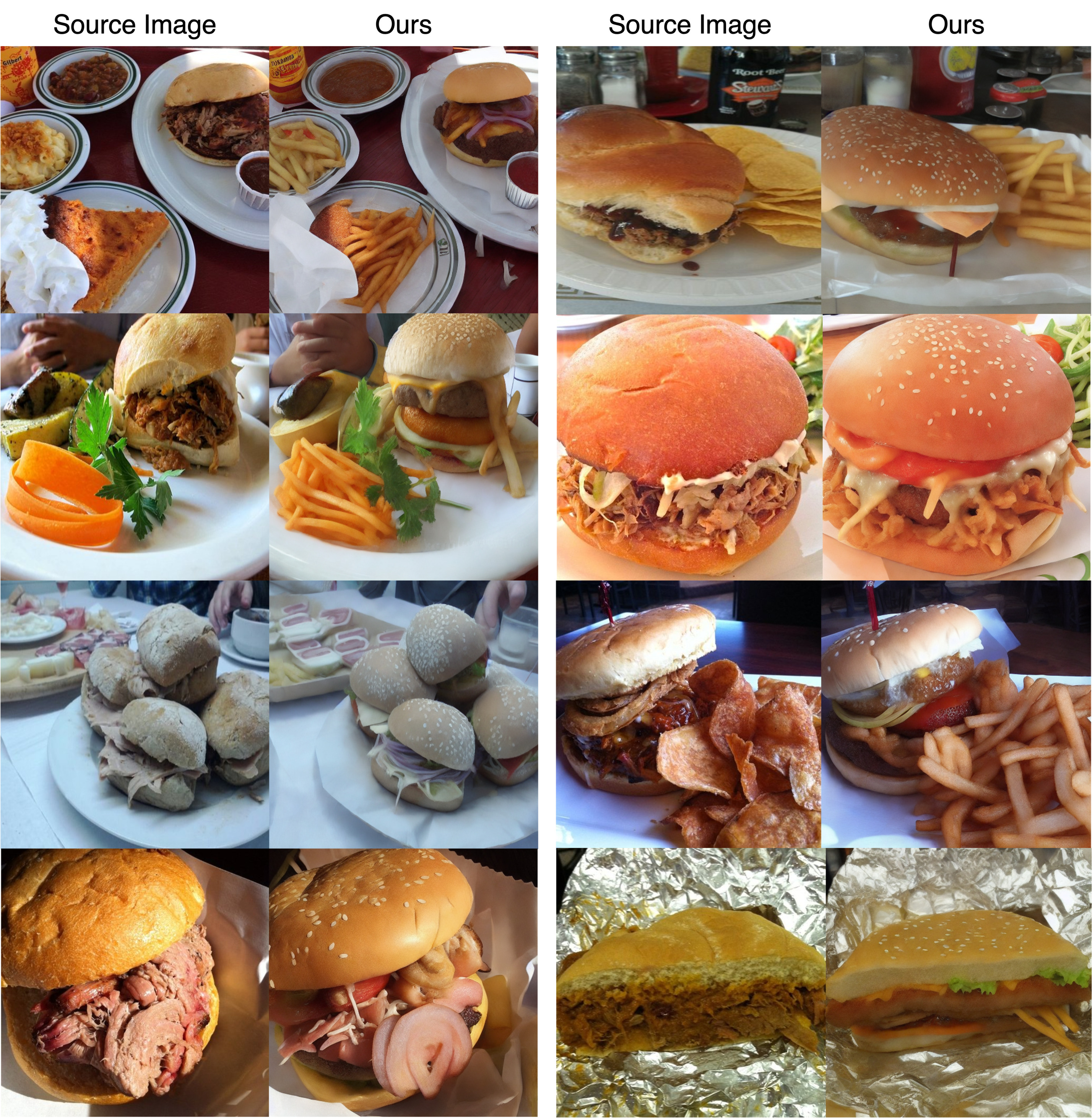}
    \caption{Image translation example by LSB for pulled-pork sandwich $\rightarrow$ hamburger task (8 NFEs). We use Fool-101 dataset.}
    \label{fig:pulledpork2hamburger}
\end{figure*}

\begin{figure*}[t]
    \centering
    \includegraphics[width=\linewidth]{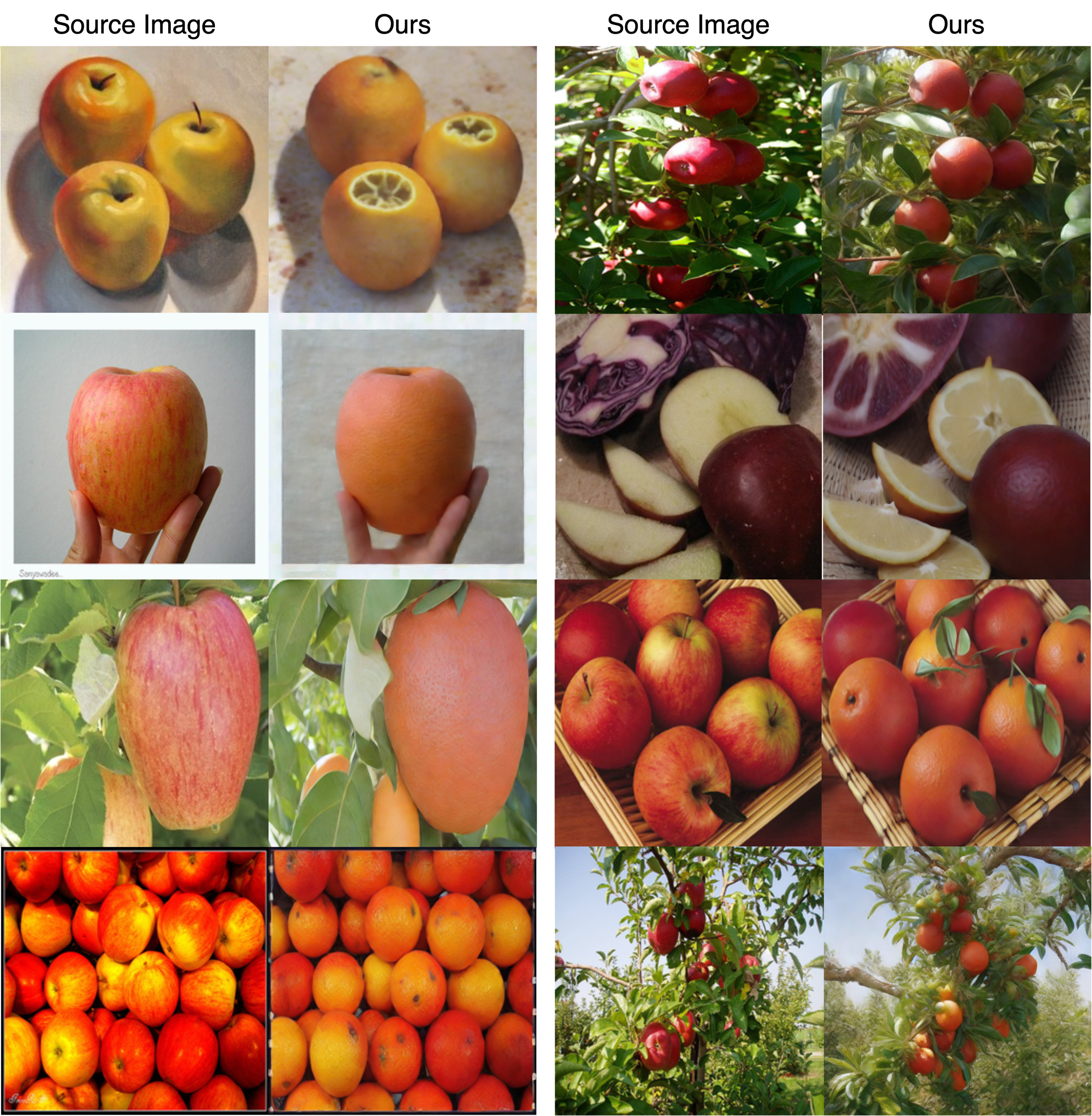}
    \caption{Image translation example by LSB for apple $\rightarrow$ orange task (8 NFEs). We use cycleGAN apple2orange dataset.}
    \label{fig:apple2orange}
\end{figure*}

The proposed method is not restricted to certain domain. To demonstrate the variety of the applicable domains for image to image translation, we illustrated additional results in Figure~\ref{fig:photo2gogh}-\ref{fig:apple2orange}. For the text optimization, we sample 1k image from training dataset and conduct textual inversion for 1k iterations. For the image translation, we use 8 NFEs, $\omega=11.0$, $t_0=0.2$ and $\sqrt\tau=0.25$.
The examples show the potential of the proposed method as a general fast image to image translation algorithm levaging Stable Diffusion, without text description for each domain. 

\section{Limitations}
Although the proposed method enables fast image-to-image translation, output quality diminishes when using 4 NFEs or fewer. This occurs because the LSB ODE is not the Schrödinger bridge as we did not construct the global OT map $\PP_{01}^\tau$; thus, training a vector field is essential for achieving higher-quality translations with lower NFEs.


\end{document}